\documentclass[10pt,twocolumn,letterpaper]{article}

\usepackage[pagenumbers]{iccv} 

\definecolor{iccvblue}{rgb}{0.21,0.49,0.74}
\usepackage[pagebackref,breaklinks,colorlinks,allcolors=iccvblue]{hyperref}

\usepackage{graphicx}
\usepackage{amsmath}
\usepackage{multirow}
\usepackage{colortbl}
\usepackage{adjustbox}
\usepackage{array}
\usepackage{booktabs}
\usepackage{makecell}
\usepackage{placeins}
\usepackage[utf8]{inputenc}
\usepackage{amssymb}
\usepackage{algorithm}
\usepackage{algpseudocode}
\usepackage{float}  

\def\paperID{9036} 
\def\confName{ICCV}
\def\confYear{2025}

\title{GVD: Guiding Video Diffusion Model for Scalable Video Distillation}

\author{
\textbf{Kunyang Li}\textsuperscript{1}\thanks{Equal contribution.},
\textbf{Jeffrey A Chan Santiago}\textsuperscript{1}\footnotemark[1],
\textbf{Sarinda Dhanesh Samarasinghe}\textsuperscript{1} \\
\textbf{Gaowen Liu}\textsuperscript{2},
\textbf{Mubarak Shah}\textsuperscript{1} \\
\textsuperscript{1}Center for Research in Computer Vision, University of Central Florida \\
\textsuperscript{2}Cisco Research, San Jose, California, USA
}

\begin{document}
\maketitle

\begin{abstract}

To address the larger computation and storage requirements associated with large video datasets, video dataset distillation aims to capture spatial and temporal information in a significantly smaller dataset, such that training on the distilled data has comparable performance to training on all of the data.
We propose \textbf{GVD: Guiding Video Diffusion}, the first diffusion-based video distillation method. 
GVD jointly distills spatial and temporal features, ensuring high-fidelity video generation across diverse actions while capturing essential motion information. 
Our method's diverse yet representative distillations significantly outperform previous state-of-the-art approaches on the MiniUCF and HMDB51 datasets across 5, 10, and 20 Instances Per Class (IPC). Specifically, our method achieves 78.29\% of the original dataset's performance using only 1.98\% of the total number of frames in MiniUCF. Additionally, it reaches 73.83\% of the performance with just 3.30\% of the frames in HMDB51.
Experimental results across benchmark video datasets demonstrate that GVD not only achieves state-of-the-art performance but can also generate higher resolution videos and higher IPC without significantly increasing computational cost.

\end{abstract}    
\section{Introduction}
\label{sec:intro}

\begin{figure}[t]
  \centering
  \includegraphics[width=\linewidth]{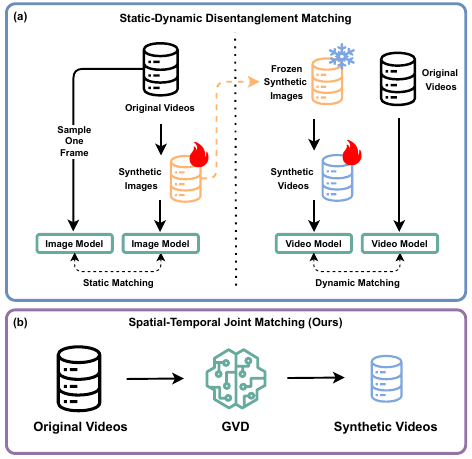} %
  \caption{
  (a) Static-Dynamic Disentanglement Matching \cite{wang2024dancing} performs static matching on images, then dynamic matching on videos. (b) Our method enhances distillation efficiency by matching spatio-temporal features with a guided video diffusion model.
  }
  \label{fig:abstract}
  \vspace{-1em}
\end{figure}

In the current data-driven era, the increasing affordability of collecting large-scale datasets has significantly accelerated the advancement of deep learning. However, training models on such massive datasets poses significant computational and storage challenges. To address this bottleneck, the distillation of datasets has emerged as a promising approach, aiming to condense large data sets into a much smaller but highly informative synthetic subset, enabling efficient model training while preserving essential knowledge from the original data \cite{yang2024dataset, shao2025context, yu2023dataset}. 

This distilled dataset should enable models to effectively distinguish between different classes by capturing representative features of each class. Additionally, to help the model learn robust and generalizable features, the distilled data should minimize redundancy and maximize sample diversity to better represent the original data distribution.
Although dataset distillation has been extensively explored in the context of images, the transition to video distillation presents rare and more complex challenges.

\begin{figure}[t]
  \centering
  \includegraphics[width=\linewidth]{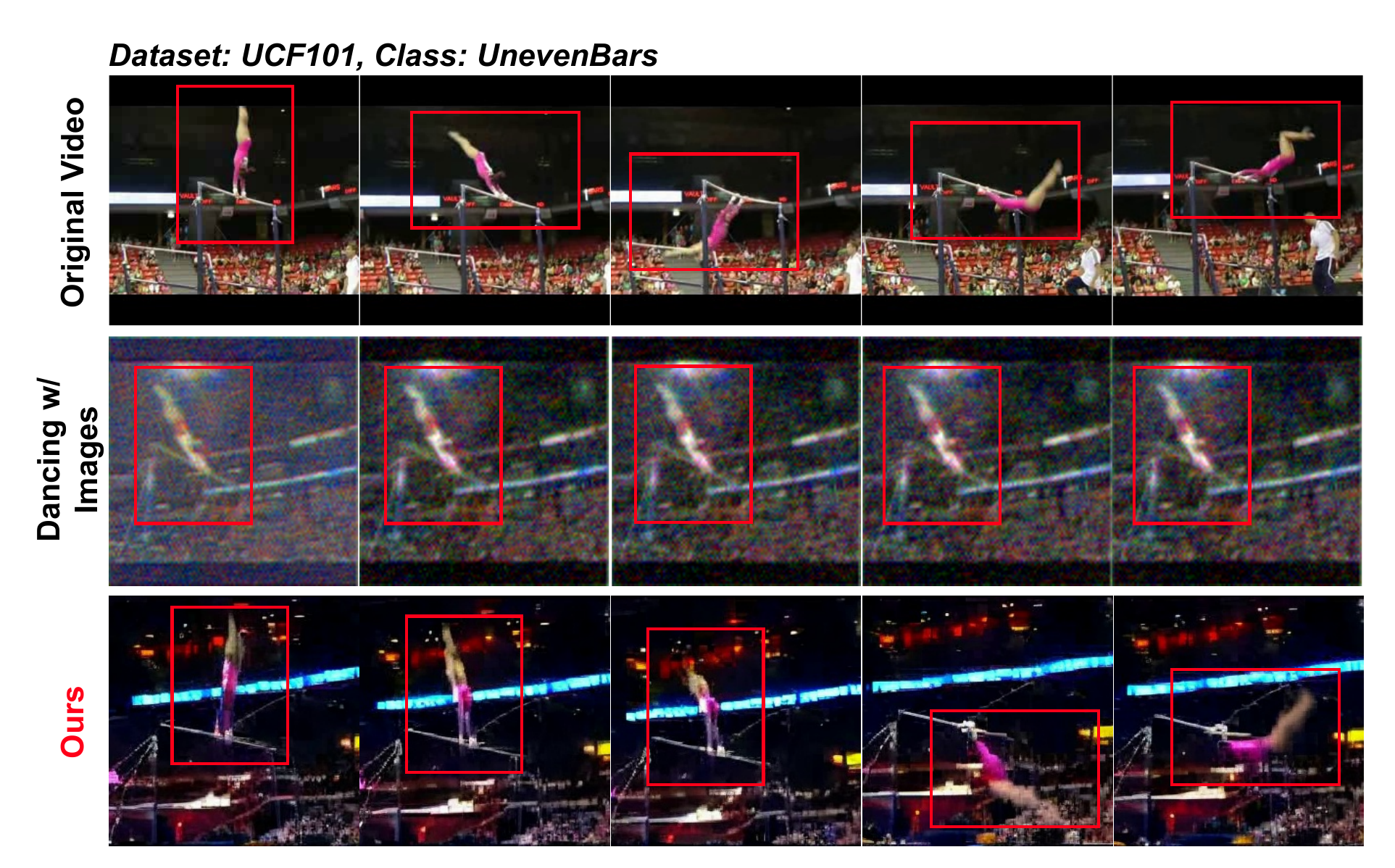} %
  \caption{We compare original and generated videos for UCF101's ``UnevenBars" class \cite{soomro2012ucf101}. Previous work \cite{wang2024dancing} (middle) captures minimal motion of the acrobat, while ours (bottom) effectively captures motion similar to the original dataset video (top).}
  \label{fig:time-lost}
  \vspace{-1.5em}
\end{figure}

Unlike static images, videos inherently exhibit large-scale complexity, as each video consists of multiple frames. For example, distilling a 16-frame video is approximately equivalent to distilling 16 images, drastically increasing the data volume and computational demand. More importantly, video data contains not only spatial information (as in images) but also temporal dynamics (motion), which is essential for capturing meaningful video representations. Traditional dataset distillation frameworks \cite{zhao2023dataset, zhao2020dataset, cazenavette2022dataset}, originally designed for images, employ matching strategies to capture critical information while generating distilled datasets via optimization in the image space. However, directly applying these paradigms to video data presents significant challenges. These methods suffer from high computational costs and struggle with large instance-per-class (IPC) and high-resolution settings. 

\begin{figure}[t]
  \centering
  \includegraphics[width=\linewidth]{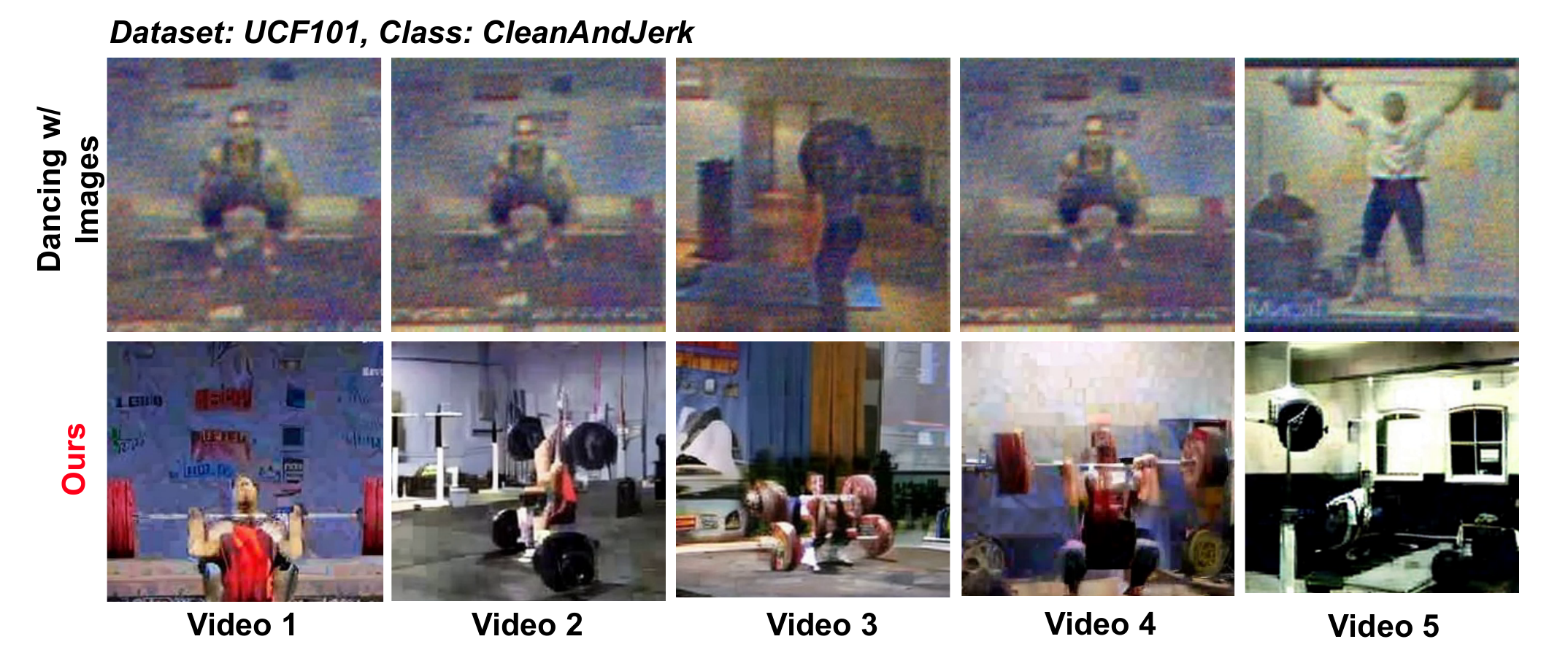} %
  \caption{
  We compare 5 generated video instances within one class, each represented by one frame. Previous work \cite{wang2024dancing} produces visually similar videos (1, 2, and 4). Our method produces higher-quality and more diverse samples. 
  } 
  \label{fig:diversity-lost}
  \vspace{-1em}
\end{figure}

Current video dataset distillation framework \cite{wang2024dancing} adopts a two-stage framework: static learning to distill spatial information and dynamic fine-tuning to compensate temporal information. Each stage synthesizes the dataset via optimization (\cref{fig:abstract}). However, these methods, which extend from image distillation frameworks, encounter several critical limitations when applied to videos: (1) \underline{\emph{Computational Complexity}}: 
when applying optimization-based approaches to video models, gradient computation and memory usage grow as a function of the number of frames, making them significantly more demanding than image-based methods. As a result, these methods become impractical for high-resolution videos or large IPC values (e.g., IPC = 10, 20).
(2) \underline{\textit{Limited Temporal Information}}: While dynamic fine-tuning aims to improve temporal representation, the generated videos do not exhibit any visually meaningful motion dynamics, as shown in \cref{fig:time-lost}.
(3) \underline{\textit{Dataset Redundancy}}: \cite{wang2024dancing} generated datasets contain repetitive samples and similar data distribution characteristics across many classes (\cref{fig:diversity-lost}).

In this work, we propose to study video dataset distillation using diffusion models, leveraging their ability to learn intrinsic data distributions and generate high-quality samples. However, directly applying standard video diffusion models \cite{wang2023modelscope} to distillation leads to a lack of diversity in the generated videos. This limitation becomes more pronounced as the number of samples per class (IPC) increases, resulting in redundant information and a significant drop in test performance.

To address this issue, we introduce the Guiding Video Diffusion Model (GVD), a novel framework that enhances intra-class diversity while maintaining motion coherence. GVD employs a Guiding Mechanism to regulate the diffusion process, preventing excessive redundancy in generated samples. Our approach integrates a Frame-wise Linear Decay Mechanism, Multi-Video Instance Composition, and soft label approach that collectively ensure smooth, coherent motion while promoting diversity—effectively distilling video data with superior representational quality and generalization.

Our key contributions can be summarized as follows:

\begin{itemize}
    \item We propose the first diffusion-based framework for video dataset distillation, introducing a novel guiding mechanism that significantly enhances intra-class diversity while preserving temporal coherence (\cref{sec:guidingdm},  \cref{fig:pipeline}).

    \item We show that our method is computationally efficient with fixed complexity regardless of IPC scale, enabling high-fidelity distillation for high-resolution videos (448×256) at large IPC values (10, 20) with minimal memory overhead (\cref{sec:exper}).

    \item Extensive experiments demonstrate that GVD achieves superior results on benchmark datasets including MiniUCF and HMDB51, reaching up to 78\% of full dataset performance while using only 1.98\% of the original frames, with robust cross-architecture generalization (\cref{sec:mainres}, \cref{sec:crossarch}).
\end{itemize}


\section{Related Work}
\label{sec:relate}

\subsection{Image Distillation}
In image distillation, recent studies classify image dataset distillation approaches into three main categories: Synthesis-Time Matching, Dual-Time Matching, and Training-Time Matching \cite{su2024d}. Early methods predominantly utilized a synthesis-time matching framework, aligning data characteristics during the generation process \cite{deng2022remember, zhao2023dataset, zhao2020dataset, cazenavette2022dataset}. Following these, SRe\textsuperscript{2}L \cite{yin2024squeeze} introduced an innovative Dual-Time Matching framework, demonstrating that decoupling bi-level optimization into distinct Squeeze, Recover, and Relabel stages can improve performance on large-scale datasets. Inspired by SRe\textsuperscript{2}L \cite{yin2024squeeze}, D\textsuperscript{4}M \cite{su2024d}, based on a Latent Diffusion Model, separates data generation and feature extraction in a compressed latent space, enabling cross-architecture generalization and broader downstream applicability. It also leverages generative diffusion models to enhance realism and diversity. 

Besides D\textsuperscript{4}M \cite{su2024d}, recent works have also leveraged diffusion models for image distillation, achieving promising results. There are \textit{Minimax Diffusion} \cite{gu2024efficient} introduces a \textit{Minimax Criterion} to improve dataset representativeness. Previous work MGD\textsuperscript{3} \cite{chan2025mgd} enables efficient synthesis of diverse distilled data through a tuning-free approach. While \cite{chan2025mgd} is designed for image distillation, this work extends its principles to video data, which introduces unique challenges such as temporal consistency and frame-wise variation. Unlike \cite{chan2025mgd}, which applies static guidance based on image cluster centroids, our approach dynamically adjusts frame-wise guidance using a linear decay mechanism to prevent over-guidance and maintain temporal coherence. Additionally, we introduce Multi-Video Instance Composition (MVIC) to enhance diversity by synthesizing distilled videos from multiple sources while preserving natural transitions. These innovations make our method distinct from \cite{chan2025mgd} and specifically tailored to the video domain.
\subsection{Video Distillation}
In the field of video dataset distillation, recent work such as \textit{Dancing with Still Images} \cite{wang2024dancing} introduces an approach where the static and dynamic components of videos are disentangled to maximize efficiency. This method consists of two key stages: static learning and dynamic fine-tuning. In the static learning stage, a single representative frame per segment is selected across training epochs, enabling gradient matching and capturing core visual content. The dynamic fine-tuning stage then introduces a series of frames as dynamic memory, enhancing static learning with temporal coherence through techniques such as trajectory matching \cite{cazenavette2022dataset} and distribution matching \cite{zhao2023dataset}.

\subsection{Video Diffusion Models}
Recent studies have utilized diffusion models to produce realistic videos, often guided by text, which serves as an intuitive and rich source of instruction \cite{wang2023modelscope, ho2022video, ho2022imagen, zhou2022magicvideo}. 
For example, ModelScopeT2V \cite{wang2023modelscope}, with its combined use of spatio-temporal convolutions and attention modules, adeptly learns temporal dependencies between frames, resulting in videos with fluid actions and consistent content. 
Notably, there exists a strong connection between video dataset distillation and video diffusion models when it comes to maintaining temporal consistency and spatial representation.
Compared to text-to-video models, 
\cite{zhang2023i2vgen, blattmann2023stable} focus on image-to-video tasks aiming at enhancing video quality, especially within the realm of content creation. 
We found that although image-to-video models are inherently adept at maintaining domain coherence with image conditions, they demonstrate limitations in temporal feature modeling. Without the guidance of a text prompt, it becomes challenging for the model to accurately represent the class-specific motion patterns.
\begin{figure*}[t]
  \centering
  \includegraphics[width=0.9\textwidth]{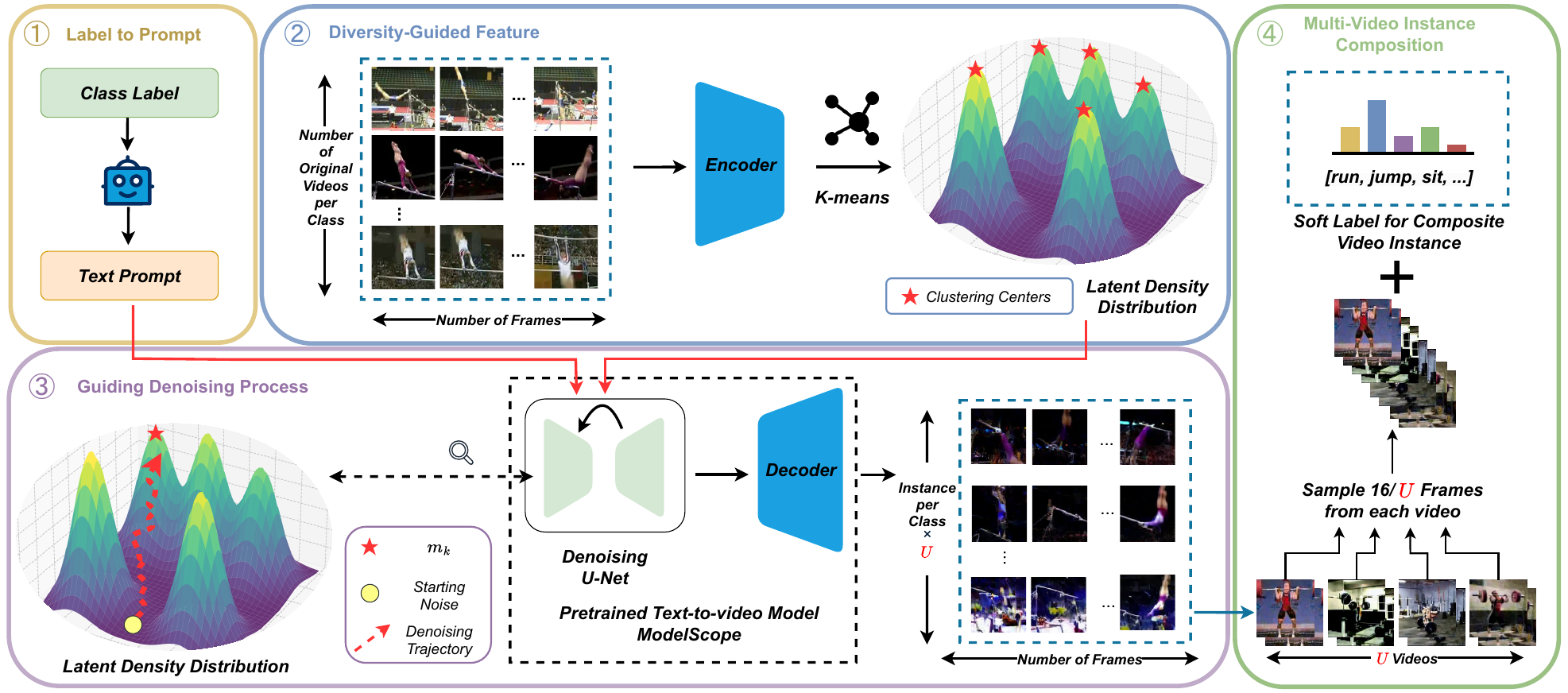} %
  \caption{ Overview of our method (GVD): (1) Generate descriptive text prompts using a language model (LLM); (2) Compute the cluster centers for each class in the real dataset; (3) Synthesize a distilled video corresponding to each class prompt and cluster center $m_k$. During the denoising process, the GVD directs the denoising trajectory toward the cluster center, represented as \( m_k \). We use \textit{ModelScope} \cite{wang2023modelscope} for distilled video synthesis. (4) We compose each instance using multiple videos by sampling and stacking their frames.}
  \label{fig:pipeline}
  \vspace{-1em}
\end{figure*}

\section{Method}
\label{sec:method}

\subsection{Preliminaries}

Due to limited computational resources and the complexity of video data matching, the previous method \cite{wang2024dancing} primarily captures spatial features and often fails to preserve temporal dynamics effectively. Our approach addresses this limitation by leveraging a pre-trained text-to-video diffusion model, which inherently encodes rich prior knowledge and robust spatio-temporal representations. 

In video diffusion, given a noisy latent representation \( Z_T \sim \mathcal{N}(0, I) \), the diffusion model gradually denoises it to generate a latent representation \( Z_0 \). At each timestep \( \hat{t} \), the U-Net predicts the current noise \( \epsilon^{\text{pr}}_{\hat{t}} \) as:
\begin{equation}
c = \tau(p),
\end{equation}
\begin{equation}
\epsilon^{\text{pr}}_{\hat{t}} = \epsilon_{\theta}(Z_{\hat{t}}, c, \hat{t}),
\end{equation}
where \( p \) is the prompt or class description, \( c \) is the text embedding obtained from \( p \), and \( Z_{\hat{t}} \) is the latent representation at timestep \( \hat{t} \). Conditioning on \( c \) allows the model to generate videos that not only match the visual appearance of the target class but also adhere to higher-level semantic attributes encoded in the text prompt. This is particularly beneficial for high-fidelity video distillation dataset generation, where temporal coherence and semantic accuracy are critical.

Then, the U-Net is trained or fine-tuned by minimizing the discrepancy between the predicted noise \( \epsilon^{\text{pr}}_{\hat{t}} \) and the ground-truth noise \( \epsilon^{\text{gt}}_{\hat{t}} \) using the following loss function:
\begin{equation}
\mathcal{L} = \mathbb{E}_{Z_{\hat{t}}, \epsilon^{\text{gt}}_{\hat{t}} \sim \mathcal{N}(0, 1), \hat{t}} \left[ \left\| \epsilon^{\text{gt}}_{\hat{t}} - \epsilon^{\text{pr}}_{\hat{t}} \right\|_2^2 \right].
\end{equation}
Notably, by fine-tuning the U-Net within the diffusion model using our specially designed text prompts that summarize video class features on the target distillation dataset, we enable the generation of videos that align closely with the distribution and distinct classes of the target data domain.

\subsection{Preliminary Experiment with K-Noise}

\begin{figure}[ht]
  \centering
  \includegraphics[width=\linewidth]{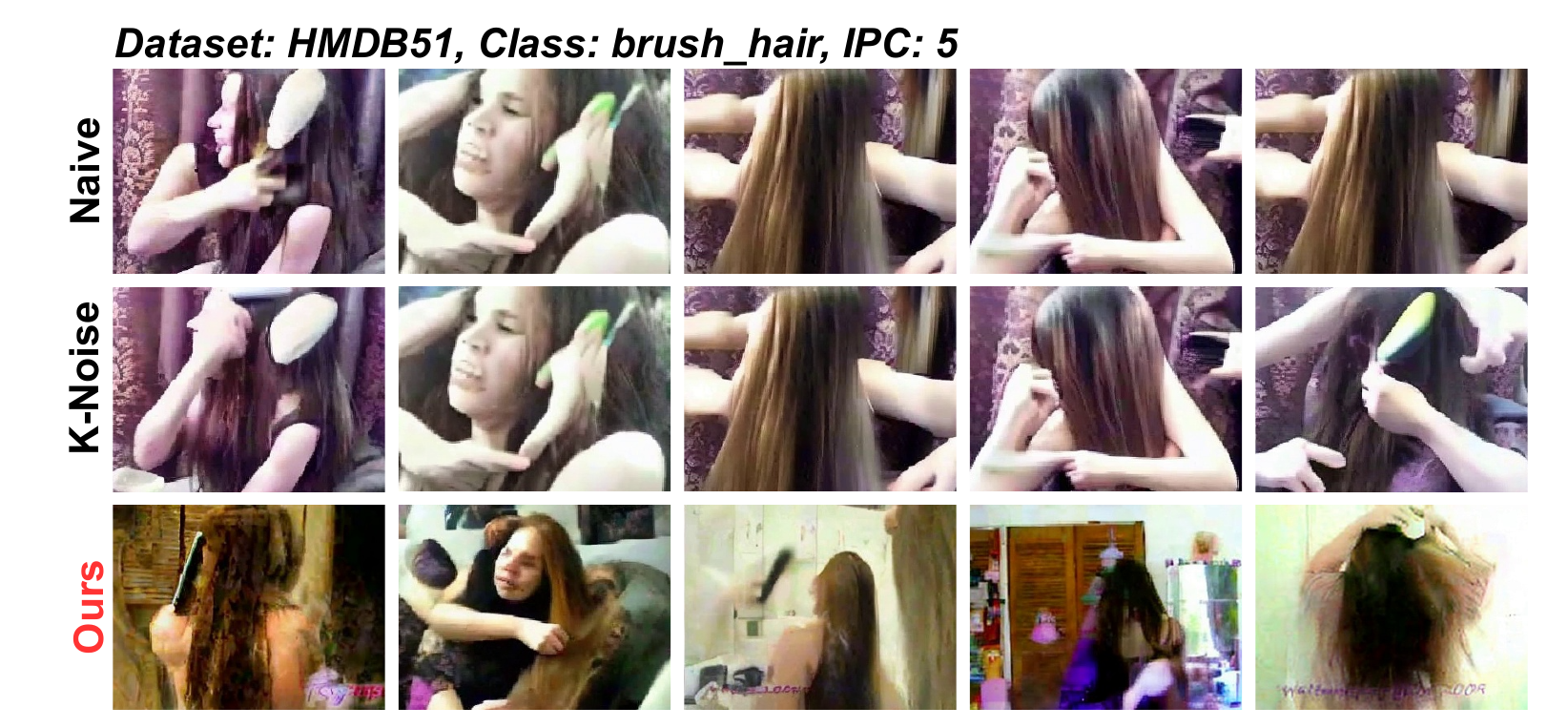} %
  \caption{A comparison of generated videos with 5 IPC across the diffusion-based techniques. 1) \textbf{Naive} uses random noise as the starting point, 2) \textbf{K-Noise}\cite{su2024d} uses noisy k-means centers as starting points. The K-noise method produces very similar videos to Naive. 3) \textbf{Our} method using GVD generates a diverse set of samples for each class.} 
  \label{fig:knoise}
    \vspace{-1em}  
\end{figure}

Inspired by D\textsuperscript{4}M \cite{su2024d}, we initially adopted cluster centers \cite{nichol2021improved} (prototypes) as the starting points for the diffusion process, treating these prototypes as representative features of each class. This strategy aims to inject dense feature information, reduce data redundancy, and improve the quality of distilled data \cite{su2024d}. However, we observed that the noise introduced in the early steps of the diffusion process destroys most of the prototype information, generating videos very similar to when using random noise as shown in \cref{fig:knoise} for the brush\_hair class. In some cases, the effect of these prototypes also diminishes diversity across class samples. 

\subsection{Guiding Video Diffusion Model}
\label{sec:guidingdm}

\cref{fig:pipeline} shows an overview of our method. Our approach utilizes cluster centers as feature-guiding vectors in the latent space throughout the denoising process rather than directly initializing with them, similarly to \cite{chan2025mgd}. By guiding the diffusion process at each step rather than starting from a noisy cluster center, we reduce the impact of early-stage noise, preventing the loss of the class prototype information. 

However, the spatio-temporal k-means centroid only represents one specific frame-to-frame transition path. In other words, it captures a single ``trajectory'' of image transformations, which does not cover the variety of ways frames can evolve in a video. Given that the generation process starts with an initial frame $f_0$ and follows a probability distribution $p(f_i | f_{<i})$ to generate subsequent frames, the clustering centroid represents a specific path that might not align well with different initial frames or transitions that could occur in other video instances. To address this limitation, we propose a mechanism for controlling frame-wise guidance, adapting \cite{chan2025mgd} for Video Distillation.


\subsubsection{Guiding Denoising Process}

Given a set of videos, each belonging to a class \( c \in \mathcal{C} \), where \( \mathcal{C} \) denotes the set of all classes, let \( v^{gt}_c \) represent a video from class \( c \) with \( F \) frames. Each frame \( f_i \) of the video \( v^{gt}_c \) is encoded into a latent representation via a VQGAN encoder \cite{esser2021taming} \( \mathcal{E} \) as follows:
\begin{equation}
    Z^{gt}_{0,c} = \left[ \mathcal{E}(f_1), \mathcal{E}(f_2), \dots, \mathcal{E}(f_F) \right],
\end{equation}
where \( Z^{gt}_{0,c} \in \mathbb{R}^{F \times \frac{H}{8} \times \frac{W}{8} \times 4} \) represents the latent variable of the video \( v^{gt}_c \) in the compressed latent space. Here, \( H \) and \( W \) denote the height and width of each frame in the video, respectively.

To generate feature-guiding vectors, we perform clustering on the latent space representations of videos within each class. For class \( c \), let \( \{ Z^{gt, i}_{0,c} \}_{i=1}^{N_c} \) denote the set of latent representations of all \( N_c \) training videos. 
For each class \( c \), we apply KMeans clustering \cite{ahmed2020k} to the set \( \{ Z^{gt, i}_{0,c} \} \) to obtain \( K \) cluster centers, where \( K \) is the number of distilled samples per class. These cluster centers serve as representative features, capturing the distribution of latent features for class \( c \) in the compressed latent space.

Let \( \{ m_{k,c} \}_{k=1}^{K} \) represent the resulting cluster centers for class \( c \):
\begin{equation}
    m_{k,c} = \text{KMeans}(\{ Z^{gt, i}_{0,c} \}_{i=1}^{N_c}), \quad k = 1, \dots, K.
\end{equation}
After getting the cluster centers, the guiding process is implemented by introducing a guiding term during generation, progressively steering the generated sample towards the target class features. This guiding term is based on the discrepancy between the target feature \( m_k \) and the current generated latent representation \( \hat{Z}_{\hat{t}} \). By incorporating a guidance coefficient \( \lambda \), the model is influenced not only by the predicted noise \( \epsilon^{\text{pr}}_{\hat{t}} \) at each denoising step but also by the direction provided by feature guidance. 

The guidance term \( g_{\hat{t}} \) is defined as the difference between \( m_k \) and the denoised prediction \( \hat{x}_0^{\hat{t}} \):
\begin{equation}
g_{\hat{t}} = m_k - \hat{x}_0^{\hat{t}},
\end{equation}
where \( \hat{x}_0^{\hat{t}} \) is the denoised prediction at timestep \( {\hat{t}} \). The guided noise prediction \( \epsilon'_{\hat{t}} \) is modified by applying a guidance strength \( \lambda \):
\begin{equation}
\epsilon'_{\hat{t}} = \epsilon_{\theta}(Z_{\hat{t}}, c, {\hat{t}}) - \lambda \cdot \sqrt{1 - \bar{\alpha}_{\hat{t}}} \cdot g_{\hat{t}}.
\end{equation}
where \( \bar{\alpha}_{\hat{t}} \) is a scaling coefficient for timestep \( {\hat{t}} \), and \( \epsilon_{\theta} \) represents the noise predicted by the model.

Since the guidance term \( g_{\hat{t}} \) is dynamically derived from the current timestep's prediction, the generation process maintains a smooth transition, effectively reducing fluctuations in the quality of generated samples. This approach ensures that the generated sample converges toward the target features without introducing abrupt changes, promoting stable and consistent sample quality throughout the generation process.

\subsubsection{Frame-wise Linear Decay Mechanism}
Excessive guidance in video generation can introduce noise. \cite{chan2025mgd} mitigates this with stop guidance, applying guidance until timestep \( t_{\text{stop}} \), after which diffusion refines the output independently. However, this alone did not fully eliminate noise (see supplemental material).  

Noise arises from a mismatch between the generated video and class prototype \( m_k \). To address this, we introduce a frame-specific guidance coefficient \( \lambda_f \) that decreases as the frame index \( f \) increases. This ensures strong guidance in initial frames while later frames rely more on preceding ones, enhancing temporal coherence. Let \( F \) denote the total number of frames.  
Then, the frame-specific guidance coefficient \( \lambda_f \) is defined as:
\begin{equation}
\lambda_f = \lambda \cdot \left(1 - \frac{f}{F}\right).
\end{equation}
Thus, the modified noise prediction for frame \( f \) becomes:
\begin{equation}
\small
\epsilon'_{{\hat{t}}, f} = 
\begin{cases}
\epsilon_{\theta}(Z_{{\hat{t}}, f}, c, {\hat{t}}) 
- \lambda_f \sqrt{1 - \bar{\alpha}_{\hat{t}}} g_{{\hat{t}}, f}, 
& \text{if } {\hat{t}} < t_{\text{stop}}, \\
\epsilon_{\theta}(Z_{{\hat{t}}, f}, c, {\hat{t}}), 
& \text{if } {\hat{t}} \ge t_{\text{stop}}.
\end{cases}
\end{equation}
As a result, the model achieves a harmony between content specificity and the natural realism of the generation, effectively mitigating issues associated with over-guidance. Finally, the latent representation \( Z_{{\hat{t}}-1, f} \) is updated for each frame \( f \) and each timestep \( {\hat{t}} \) as follows:

\begin{equation}
Z_{{\hat{t}}-1, f} = \sqrt{\bar{\alpha}_{{\hat{t}}-1}} \hat{x}_0^{\hat{t}} + \sqrt{1 - \bar{\alpha}_{{\hat{t}}-1}} \epsilon'_{{\hat{t}}, f}.
\end{equation}

\subsection{Multi-Video Instance Composition}

\begin{table*}[t]
    \centering
    \caption{Top-1 accuracy (\%) with standard deviation on the miniUCF and HMDB51 datasets using the MiniC3D network, with varying IPC. Results for \textit{Direct Image Distillation Adaptation} and \textit{Dancing with Still Images} are reported from paper \cite{wang2024dancing}. + D denotes Dynamic\cite{wang2024dancing}.}
    \renewcommand{\arraystretch}{1.2} 
    \setlength{\tabcolsep}{4pt} 
    \begin{adjustbox}{width=0.9\linewidth} 
    \begin{tabular}{ll|cccc|cccc} 
        \toprule
        \multicolumn{2}{c}{\multirow{2.5}{*}{\textbf{Method}}} & \multicolumn{4}{|c|}{\bf MiniUCF} & \multicolumn{4}{c}{\bf HMDB51} \\ 
        \cmidrule(lr){3-6} \cmidrule(lr){7-10}
        & & IPC=1 & IPC=5 & IPC=10 & IPC=20 & IPC=1 & IPC=5 & IPC=10 & IPC=20 \\ 
        \midrule
        
        \multirow{3}{*}{\makecell[l]{Coreset\\Selection}} & Random & $10.3 \pm 1.1$ & $20.7 \pm 0.5$ & $26.0 \pm 0.6$ & $35.7 \pm 0.2$ & $3.8 \pm 0.4$ & $6.5 \pm 0.5$ & $9.0 \pm 0.5$ & $12.6 \pm 1.0$ \\ 
         & Herding \cite{welling2009herding} & $11.4 \pm 0.3$ & $22.3 \pm 0.1$ & $28.7 \pm 0.9$ & $35.7 \pm 0.5$ & $4.1 \pm 0.3$ & $8.5 \pm 0.5$ & $10.6 \pm 0.2$ & $13.2 \pm 0.6$ \\
         & K-center \cite{sener2017active}& $10.0 \pm 1.5$ & $19.2 \pm 0.0$ & $26.0 \pm 0.8$ & $33.1 \pm 1.0$ & $3.7 \pm 0.7$ & $6.9 \pm 0.1$ & $7.4 \pm 0.1$ & $10.5 \pm 0.9$ \\ \midrule
        
        \multirow{4}{*}{\makecell[l]{Direct\\Image Distillation\\Adaptation}} & DM \cite{zhao2023dataset} & $15.3 \pm 1.1$ & $25.7 \pm 0.2$ & - & - & $6.1 \pm 0.2$ & $8.0 \pm 0.2$ & - & - \\ 
         & MTT \cite{cazenavette2022dataset} & $19.0 \pm 0.1$ & $28.4 \pm 0.7$ & - & - & $6.6 \pm 0.5$ & $8.4 \pm 0.6$ & - & - \\ 
         & FrePo \cite{zhou2022dataset} & $20.3 \pm 0.5$ & $30.2 \pm 1.7$ & - & - & $7.2 \pm 0.8$ & $9.6 \pm 0.7$ & - & - \\ 
         & Static-DC \cite{zhao2020dataset} & $13.7 \pm 1.1$ & $24.7 \pm 0.5$ & - & - & $5.1 \pm 0.9$ & $7.8 \pm 0.4$ & - & - \\ \midrule
        
        \multirow{3}{*}{\makecell[l]{Dancing with\\Still Images}} & DM + D \cite{wang2024dancing} & $17.5 \pm 0.1$ & $27.2 \pm 0.4$ & - & - & $6.0 \pm 0.4$ & $8.2 \pm 0.1$ & - & - \\ 
         & MTT + D  \cite{wang2024dancing} & $\mathbf{23.3 \pm 0.6}$ & $28.3 \pm 0.0$ & - & - & $6.5 \pm 0.1$ & $8.9 \pm 0.6$ & - & - \\ 
         & FRePo + D \cite{wang2024dancing} & $22.0 \pm 1.0$ & $31.2 \pm 0.7$ & - & - & $8.6 \pm 0.5$ & $10.3 \pm 0.6$ & - & - \\ \midrule
        
        \multirow{2}{*}{\textbf{GVD}} & \textbf{Ours} & $20.1 \pm 0.2$ & $\mathbf{33.3 \pm 0.8}$ & $\mathbf{39.4 \pm 0.3}$ & $\mathbf{42.1 \pm 0.1}$ & $7.2 \pm 0.2$ & $\mathbf{13.7 \pm 0.9}$ & $\mathbf{15.6 \pm 0.5}$ & $\mathbf{17.9 \pm 0.6}$ \\ 

        & \textbf{Ours w/ Soft-Label} & $22.9 \pm 0.3$ & $\mathbf{34.7 \pm 0.3}$ & $\mathbf{39.8 \pm 0.2}$ & $\mathbf{44.8 \pm 0.4}$ & $\mathbf{9.5 \pm 0.2}$ & $\mathbf{14.3 \pm 0.2}$ & $\mathbf{17.7 \pm 0.1}$ & $\mathbf{21.1 \pm 0.9}$ \\

         \cmidrule(lr){1-2} \cmidrule(lr){3-6} \cmidrule(lr){7-10}
        
        \multicolumn{2}{c}{Full Dataset} & \multicolumn{4}{|c|}{$57.2 \pm 0.14$} & \multicolumn{4}{c}{$28.6 \pm 0.69$} \\ 
        \bottomrule
    \end{tabular}
    \end{adjustbox}
    \label{tab:miniucf_hmdb51}
    \vspace{-0.5em}
\end{table*}




To maximize information density within the limited dataset size, we apply Multi-Video Instance Composition (MVIC), which constructs new video instances by sequentially selecting frames from multiple video instances of the same class. This approach enhances diversity and ensures each distilled sample encapsulates richer essential information, improving dataset effectiveness.  

To construct the final distilled dataset, we first generate \( U \times \text{IPC} \) 16-frame video instances using our guiding mechanism. The distilled videos for each class are then divided into \( U \) non-overlapping groups, where we sequentially extract \( N \) frames (\( N = 16/U \) in our experiments) from each video, ensuring the preservation of important temporal information.

We then concatenate these selected frames from all videos within the group to form a new distilled video. This ensures that each distilled video integrates meaningful and diverse content while maintaining natural temporal coherence. 

\subsection{Label Softening for Video Distillation}
Previous research \cite{muller2019does, su2024d, sun2024diversity} shows that soft labels enhance data distillation by providing richer supervision than one-hot labels, addressing inaccuracies from noisy annotations. This leads to improved model robustness and generalization. Therefore, we implement soft label training to better capture nuanced patterns.




Let \( T(x) \) represent the soft label predictions from the pre-trained teacher network for an input video \( x \), and \( S_\theta(x) \) represent the predictions from the student model with parameters \( \theta \). To encourage the student model to align with the teacher's soft label distribution, we minimize the Kullback-Leibler (KL) divergence:
\begin{equation}
\theta_{\text{student}} = \arg \min_{\theta \in \Theta} L_{KL}(\alpha T(x) + (1 - \alpha) y , S_\theta(x)),
\end{equation}
where \( y \) is a one-hot label vector, \( \alpha \) is a scalar between 0 and 1, and \( L_{KL} \) quantifies the difference between the teacher and student distributions. Both the teacher and student predictions are scaled by a temperature factor to control the smoothness of the output distributions.

\section{Experiments}
\label{sec:exper}

\subsection{Setting and Evaluation}

In this section, we describe the experimental details and evaluate the performance of our method across multiple datasets and network architectures.  

\textbf{Distillation Setup.} During distillation, we use ModelScope (Text-to-Video) \cite{wang2023modelscope} as the base model, converting class labels into text prompts for conditioning. Generation runs for 2000 timesteps, stopping guidance at step 1000. Videos are generated at 448×256 resolution, then resized for fair comparison.

\textbf{Evaluation Protocol.} For evaluation, we follow the evaluation protocol established by \cite{wang2024dancing}, utilizing one of the following lightweight networks: CNN-GRU, CNN-RNN, or MiniC3D. Experiments are conducted on NVIDIA A100 80GB GPUs. 
we trained the evaluation network on the distilled dataset with 500 epochs, a batch size of 128, and a learning rate of 0.01, which is decayed by a factor of 0.1 at epoch 250, and the SGD optimizer, primarily on the MiniC3D network. The networks are trained from scratch on the distilled datasets and evaluated on the full test set. For MiniUCF and HMDB51, videos are uniformly sampled to 16 frames, each cropped and resized to 112x112, with a 50\% random horizontal flip applied during training. We report Top-1 accuracy for MiniUCF and HMDB51. 


\textbf{Datasets.} We used several key action recognition datasets. The MiniUCF dataset, a smaller version of the original dataset \cite{soomro2012ucf101}, contains 6,523 videos across 50 action classes \cite{wang2024dancing}. We also used the HMDB51 dataset \cite{kuehne2011hmdb}, which has 6,766 videos across 51 categories, offering unbiased performance evaluation. 

\subsection{Video Distillation Results}
\label{sec:mainres}

For video distillation evaluation, we compare our method against:
\begin{itemize}
    \item Coreset Selection: Herding \cite{welling2009herding}, k-center \cite{sener2017active}, and random selection.
    \item Direct Image Distillation Adaptation: Image distillation methods (DC \cite{zhao2020dataset}, DM \cite{zhao2023dataset}, MTT \cite{cazenavette2022dataset}) applied to video, following \textit{Dancing with Still Images} \cite{wang2024dancing}.
    \item Video Distillation: \textit{Dancing with Still Images} \cite{wang2024dancing}, using dynamic fine-tuning with DM, MTT, and FRePo, denoted as DM+D, MTT+D, and FRePo+D.
\end{itemize}


\begin{figure*}[t]
  \centering
  \includegraphics[width=1.0\textwidth]{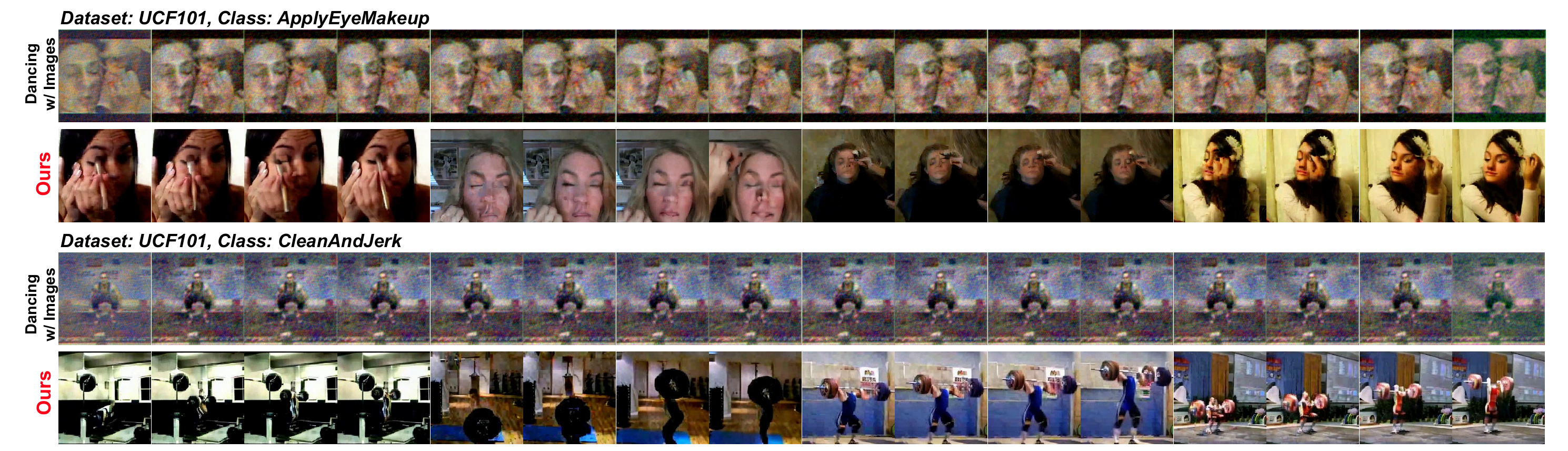} %
  \caption{We qualitatively compare our method to previous work \cite{wang2024dancing}. Our generated videos are higher resolution and capture essential and more diverse spatial-temporal information when compared to previous SOTA.}
  \label{fig:fullwidth}
  \vspace{-1em}
\end{figure*}
\textbf{MiniUCF.}
Our method achieves the highest Top-1 accuracy at IPC levels of 5, 10, and 20, with absolute improvements over previous baselines of 3.5, 11.1, and 9.1, respectively. Notably, MTT performs slightly better than MTT+D at IPC = 5, suggesting that MTT+D relies heavily on image-based distillation. 
Our method underperforms at IPC 1, which is expected since it aligns with the data distribution. At this level, a generative model can at best produce the optimal example.
Both Direct Image Distillation and Dancing with Images do not scale well to higher IPCs, such as 10 and 20. As shown in Table \ref{tab:miniucf_hmdb51}, our method achieves nearly 78.29\% of the original dataset’s performance using only 1.98\% of the frames (from 807013 in the original training set), demonstrating its efficiency and practicality in video distillation.

\textbf{HMDB51.}
Our method outperformed the others and achieved the highest Top-1 accuracy across all IPCs, with improvements of 0.9, 4.0, 7.1, and 7.9, respectively. MTT+D and DM+D performed similarly to herding at IPC = 5. Our approach also surpassed Dancing with Still Images \cite{wang2024dancing} by about 4\%, highlighting its effectiveness in distilling spatiotemporal features. At IPC = 20, it achieved $73.83\%$ of the full dataset’s training performance using only 3.3\% of the frames.

\subsection{Cross-Architecture Generalization}
\label{sec:crossarch}


\begin{table}[t]
    \centering
    \caption{Comparison of distillation performance across different architectures (ConvNet3D, CNN+GRU, and CNN+RNN) and IPC values on MiniUCF. The results are without soft labels.}
    \renewcommand{\arraystretch}{1.2}
    \resizebox{\linewidth}{!}{
    \begin{tabular}{c c c c c} 
        \toprule
        \multirow{1}{*}{\bf IPC} & \multirow{1}{*}{\bf Method} & \multicolumn{1}{c}{\bf ConvNet3D} & \multicolumn{1}{c}{\bf CNN+GRU} & \multicolumn{1}{c}{\bf CNN+RNN} \\ 
        \midrule
        \multirow{3}{*}{1} & Random & $10.3 \pm 1.1\%$ & $9.5 \pm 0.4\%$ & $12.6 \pm 0.4\%$ \\ 
        & Dancing w/ Images & $\mathbf{23.3 \pm 0.6\%}$ & $14.8 \pm 0.1\%$ & $14.2 \pm 0.2\%$ \\ 
        & \textbf{Ours} & $20.1 \pm 0.2\%$ & $\mathbf{19.3 \pm 0.5\%}$ & $\mathbf{20.1 \pm 0.1\%}$ \\ 
        \midrule
        \multirow{3}{*}{5} & Random & $20.7 \pm 0.5\%$ & $18.8 \pm 1.5\%$ & $16.8 \pm 0.0\%$ \\ 
        & Dancing w/ Images & $28.3 \pm 0.0\%$ & $15.4 \pm 0.6\%$ & $14.8 \pm 0.9\%$ \\ 
        & \textbf{Ours} & $\mathbf{33.3 \pm 0.8\%}$ & $\mathbf{29.0 \pm 0.9\%}$ & $\mathbf{28.8 \pm 0.1\%}$ \\ 
        \bottomrule
    \end{tabular}
    }
    \label{tab:miniUCF101_cross_architecture}
    \vspace{-1em}
\end{table}
\cref{tab:miniUCF101_cross_architecture} shows a significant performance drop in MTT+D when transferring from MiniC3D to CNN+GRU and CNN+RNN, indicating that dataset optimization methods lack cross-architecture generalization and introduce bias in distilled data. In contrast, our method remains stable across architectures, demonstrating greater adaptability. Additionally, our approach performs better with the more temporally focused CNN+GRU and CNN+RNN model, highlighting its effectiveness in distilling temporal features.

\subsection{Ablation Study and Analysis}
In this section, we systematically analyze the role of each component within our framework, highlighting their contributions to performance. We also examine the effectiveness of our distilled video generation approach, demonstrating its superiority in terms of efficiency, diversity, and representativeness. 

\begin{figure*}[t]
  \centering
  \begin{subfigure}{0.255\linewidth}
    \includegraphics[width=\linewidth]{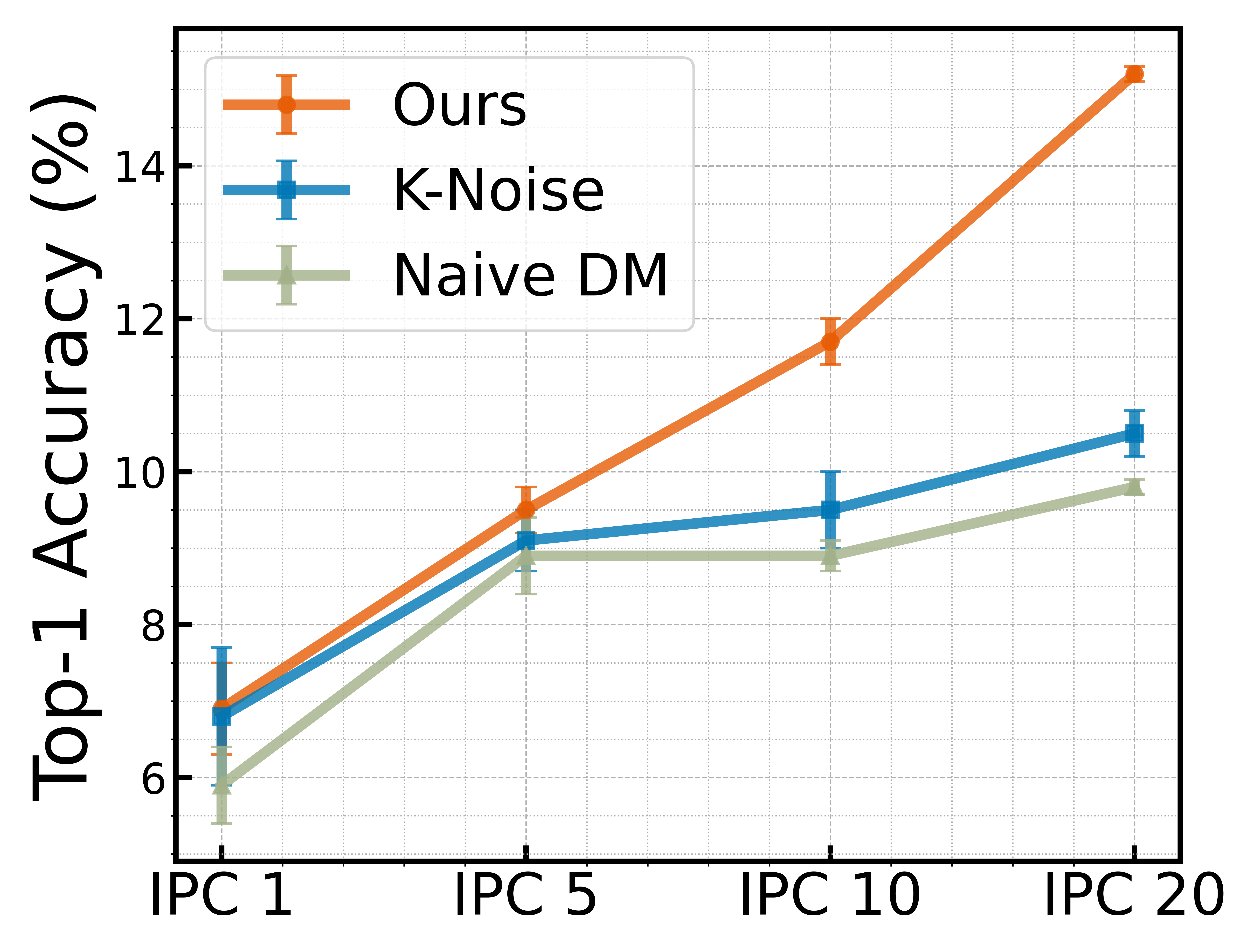}
    \caption{}
  \end{subfigure}
  \begin{subfigure}{0.22\linewidth}
    \includegraphics[width=\linewidth]{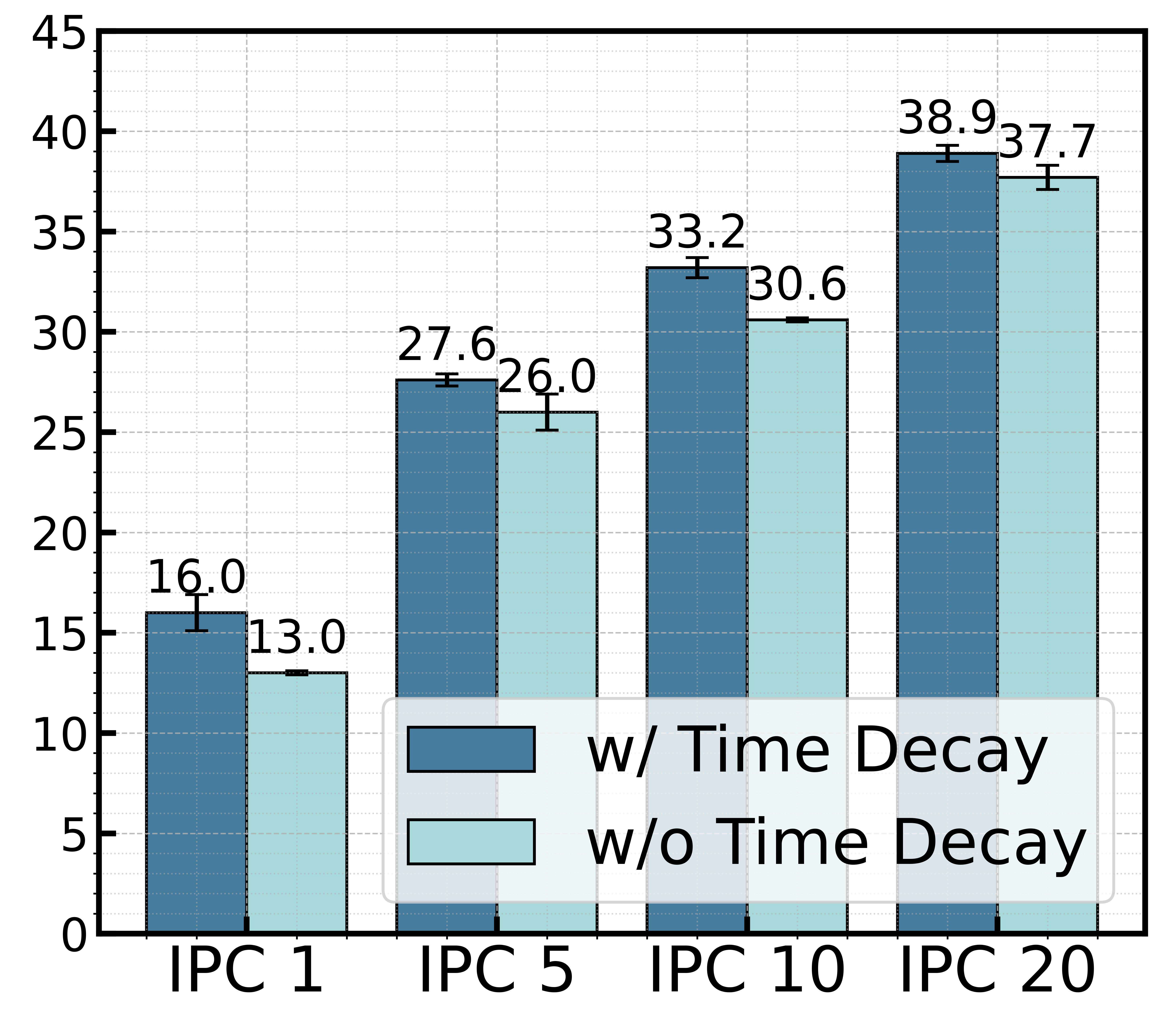}
    \caption{}
  \end{subfigure}
  \vspace{0.3cm}
  \begin{subfigure}{0.22\linewidth}
    \includegraphics[width=\linewidth]{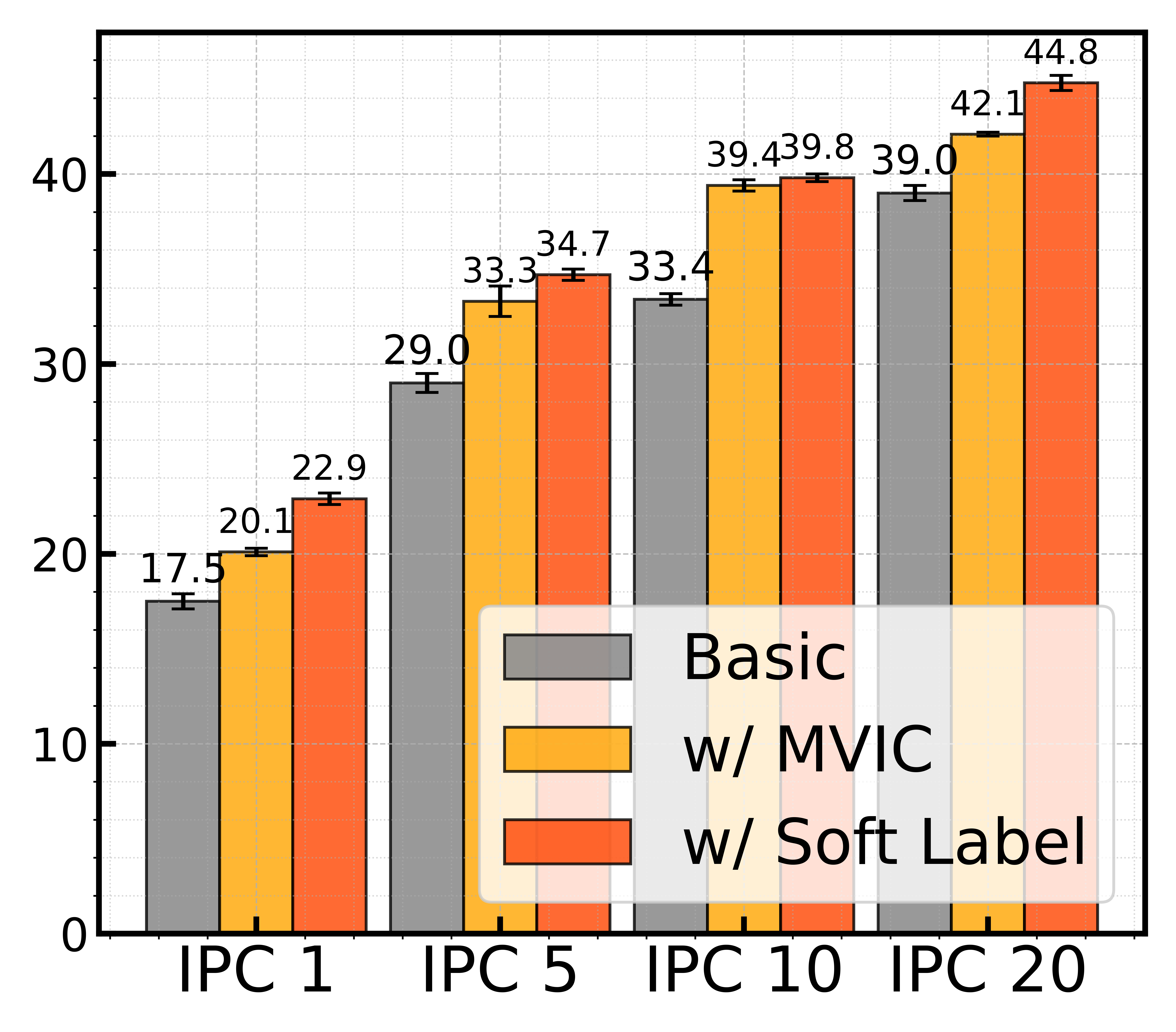}
    \caption{}
  \end{subfigure}
  \begin{subfigure}{0.22\linewidth}
    \includegraphics[width=\linewidth]{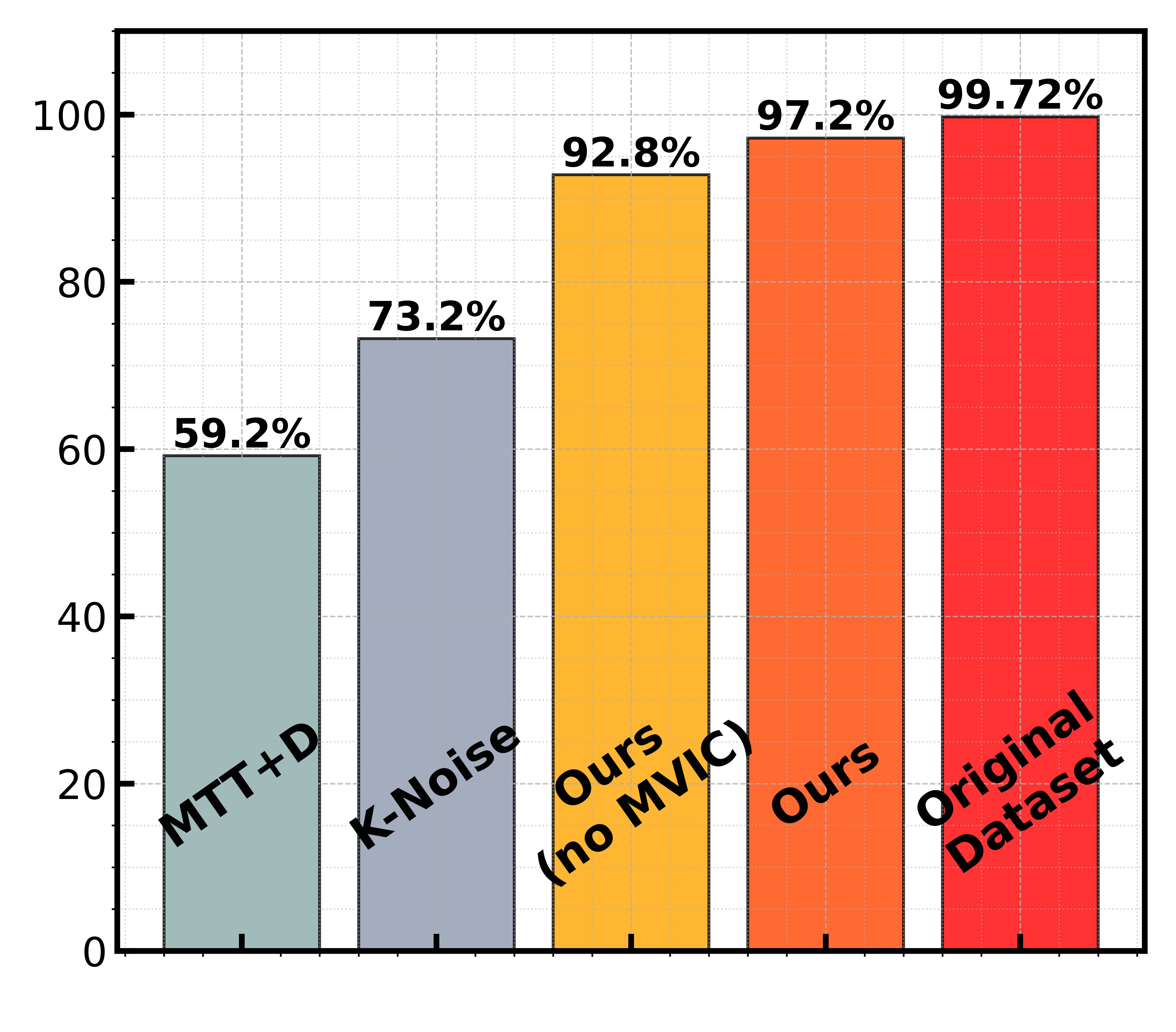}
    \caption{}
  \end{subfigure}
  \vspace{-1.5em}
  \caption{
  (a) Analysis of video diffusion sampling techniques on HMDB51 \cite{kuehne2011hmdb} distillation. 
  (b) Ablation study of frame-wise linear decay on miniUCF \cite{soomro2012ucf101}. 
  (c) Ablation study of soft labeling and video composition on miniUCF.
  (d) Representativeness analysis using pre-trained miniC3D classification accuracy of the distilled dataset.
  }
  \label{fig:ablation_study}
  \vspace{-1em}
\end{figure*}

\subsubsection{Diffusion Sampling Strategies Analysis}
In this part, we compare three methods for performing video dataset distillation using a video diffusion model:
\begin{itemize}
    \item \textbf{Naive DM}: Fine-tunes the VDM on the target dataset and generates videos per class using text prompts.
    \item \textbf{K-Noise}: Clusters condensed features as prototypes, adds noise, and denoises them to synthesize a semantically enriched dataset. To maintain temporal coherence, clustering is based on spatial features from the first frame, following D\textsuperscript{4}M \cite{su2024d}.
    \item \textbf{Ours}: Guides each DDIM sampling step with condensed features. Using spatio-temporal features without decoupling improves alignment with dense semantic regions.
\end{itemize}
\cref{fig:ablation_study} (a) shows evaluation results on HMDB51 \cite{kuehne2011hmdb}. Our method outperforms baselines, with a growing performance gap as IPC increases, demonstrating superior condensation of spatio-temporal features. The distilled features also exhibit greater diversity and higher quality, enhancing data representation accuracy.

\subsubsection{Components Analysis}  
We analyze three key components of our method: frame-wise linear decay, MVIC, and label softening.  

\textbf{Frame-wise Linear Decay:} Fig. \ref{fig:ablation_study} (b) demonstrates that incorporating a frame-wise linear decay mechanism consistently improves performance across all IPC values (1, 5, 10, 20) by at least 1.2\%, enhancing temporal modeling for more accurate evaluations.  

\textbf{Multi-Video Instance Composition:} As shown in Fig. \ref{fig:ablation_study} (c), MVIC significantly impacts performance, which is evident from the gap between the gray (first) and orange (second) bars. To balance diversity and temporal coherence, we set \( U = 4 \), composing each distilled video from 4 randomly sampled frames per video. This preserves temporal structures while incorporating diverse features. Grid search confirms that random grouping and frame sampling yield the best results (see Appendix \ref{sec:A-9-composition} for details). On MiniUCF \cite{soomro2012ucf101}, reconstruction improves performance by 4.0\%.

\textbf{Label Softening:} Further leveraging reconstructed videos, we analyze soft-label effects in Fig. \ref{fig:ablation_study} (c). Comparing the orange (second) and orange-red (third) bars shows that soft labels from a pre-trained VideoMAE \cite{tong2022videomae} improve performance. Using output probabilities as soft labels with $\alpha$ = $0.2$ and temperature = 3.0, performance gains on MiniUCF are $2.8$, $1.4$, $0.4$, and $2.7$ for each IPC.  

\subsubsection{Efficiency Comparison}
Our method scales efficiently with stable 50.08GB memory usage across IPC values, ensuring minimal computing increase. In contrast, Dancing with Still Images \cite{wang2024dancing} exhibits linear growth in time and memory, limiting scalability.


\begin{table}[t]
    \centering
    \caption{Evaluation on diversity metrics. Our method demonstrates superior diversity across all metrics.}
    \renewcommand{\arraystretch}{1.2} 
    \setlength{\tabcolsep}{8pt} 
    \resizebox{0.9\linewidth}{!}{ 
        \begin{tabular}{lccc}
            \toprule
            \textbf{Method} & \textbf{Entropy} ($\uparrow$) & \textbf{Coverage} ($\uparrow$) & \textbf{MPD} ($\uparrow$) \\ 
            \midrule
            Dancing       & 2.1428           & 0.0215           & 0.8341  \\
            K-Noise         & 4.1576           & 0.4362           & 0.8428  \\
            \textbf{Ours}  & \textbf{4.3409}  & \textbf{0.4758}  & \textbf{0.9059} \\
            \bottomrule
        \end{tabular}
    }
    \label{tab:diversity_metrics}
    \vspace{-1em}
\end{table}
\subsubsection{Diversity and Representativeness Comparison}
We evaluate the diversity of the distilled dataset using entropy, coverage, and mean pairwise distance on a pretrained VideoMAE model on UCF101, capturing different aspects of dataset richness and variability (\cref{sec:diversitymetrics}). Our method outperforms Dancing with Images and K-Noise across all three metrics.  

For representativeness, we use a MiniC3D network pretrained on the original dataset as a scoring model on MiniUCF (IPC = 5). A more representative dataset should yield higher accuracy, preserving key class-specific information. As shown in Fig. \ref{fig:ablation_study} (d), compared to Dancing with Still Images\cite{wang2024dancing} (59.2\%) and K-Noise (73.2\%), our method achieves significantly better feature retention, approaching the original dataset's accuracy (99.72\%).
Additionally, classification accuracy improves from 92.8\% to 97.2\% after video composition, demonstrating that multiple instances composition enhances both representativeness and diversity. Our method’s strong cross-architecture generalization (Table \ref{tab:miniUCF101_cross_architecture}) further supports its effectiveness.

\section{Conclusion}
\label{sec:con}
In this work, we introduced GVD: Guiding Video Diffusion, the first video diffusion-based approach for Video Distillation following the representativeness and diversity paradigm. GVD achieves state-of-the-art performance on MiniUCF and HMDB51 with minimal computational and memory overhead. It retains over 78\% of the original accuracy on MiniUCF using less than 2.0\% of frames and 74\% on HMDB51 with 3.3\% of frames. Additionally, GVD produces a distilled dataset robust across architectures, enhancing versatility. These results establish GVD as a practical, scalable, and efficient solution for video distillation.  

\section*{Acknowledgements}
The authors would like to thank the anonymous reviewers for their insightful feedback. This work was partially supported by a research gift from Cisco.

{
    \small
    \bibliographystyle{ieeenat_fullname}
    \bibliography{main}
}

\appendix
\clearpage
\setcounter{page}{1}
\maketitlesupplementary

\section{Diversity Metrics}
\label{sec:diversitymetrics}

\textbf{Entropy:} measures the uncertainty or distributional spread of the dataset across different classes or feature representations. Higher entropy indicates a more uniform distribution of samples, meaning the dataset covers a broader range of variations rather than being concentrated in a few modes. In our case, a higher entropy suggests that the distilled dataset retains diverse feature representations, avoiding mode collapse.

\textbf{Coverage:} Coverage measures how well the distilled dataset preserves the diversity of the original dataset’s feature space. Higher coverage indicates better retention of class-specific features, ensuring that essential modes are not lost. We compute coverage as the proportion of points in \( D_{\text{orig}} \) that are sufficiently close to at least one point in \( D_{\text{small}} \). Specifically, for each \( x \in D_{\text{orig}} \), we find its nearest neighbor \( y \in D_{\text{small}} \) using Euclidean distance. A point is considered covered if:
\begin{equation}
    \text{dist}(x, y) \leq \tau,
\end{equation}
where \( \tau \) is the 90th percentile of nearest-neighbor distances in \( D_{\text{orig}} \). The coverage metric is then computed as:
\begin{equation}
    C = \frac{\left| \{ x \in D_{\text{orig}} \mid \exists y \in D_{\text{small}}, \text{dist}(x, y) \leq \tau \} \right|}{|D_{\text{orig}}|}.
\end{equation}
This quantifies how well \( D_{\text{small}} \) spans the original feature space.



\textbf{Mean Pairwise Distance (MPD):} measures the average Euclidean distance between pairs of samples in the feature space. A higher MPD suggests that the dataset contains more diverse samples, as the features are more spread out rather than clustered too closely. 

\section{Frame Condensation}

\begin{figure}[!ht]
  \centering
  \includegraphics[width=0.85\linewidth]{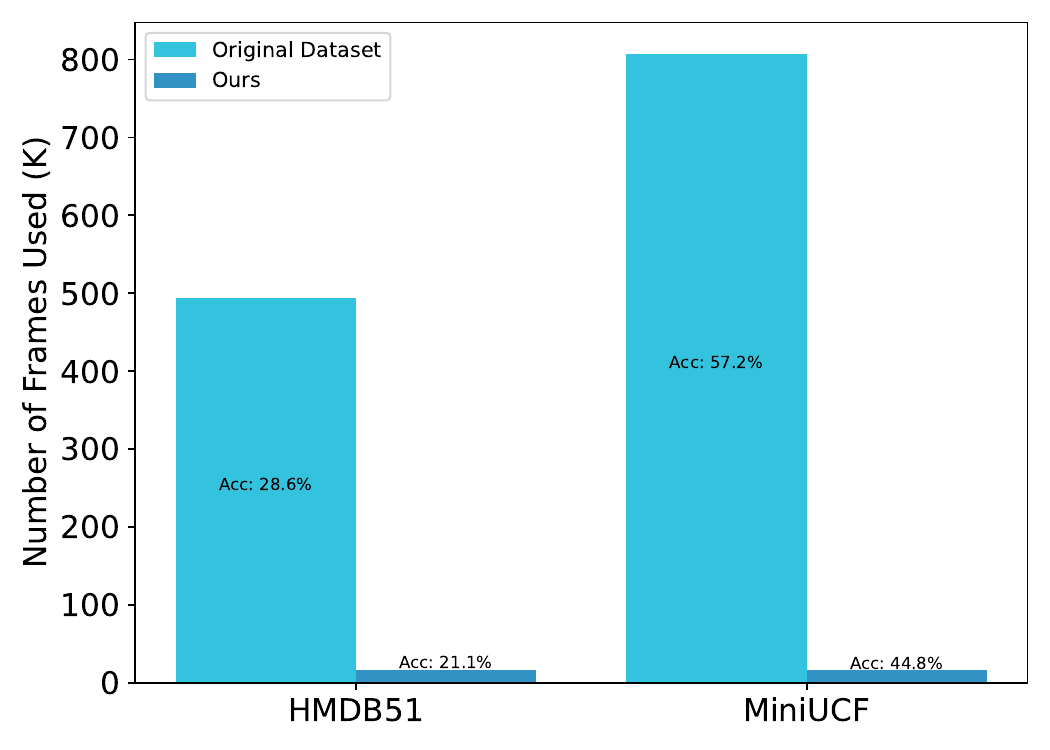} %
  \caption{Comparison of the average frames per video between the original datasets (estimated) and distillation datasets for HMDB51 and MiniUCF. This significant reduction in frames per video highlights the efficiency of the distillation process in condensing the video information.}
  \label{fig:condensation}
\end{figure}

Figure~\ref{fig:condensation} presents the comparison of average frames per video between the original datasets and their distilled counterparts for different datasets. The average frames per video in the original datasets are estimated by dividing the total number of frames by the total number of videos. In contrast, the distillation standardizes the number of frames per video to a fixed value of 16 for both datasets. While this method drastically reduces the data size, it retains critical features necessary for downstream tasks, such as video recognition. 

\label{sec:A2}

\section{Empirical Analysis of Clustering Centers}
\label{sec:A4}

In this section, we explore various clustering methods to construct the guiding term for video data. All clustering is conducted in the encoded feature space \cite{esser2021taming}. Since video data is inherently 3D, we extend three methods based on the 2D k-means \cite{ahmed2020k} algorithm:

\begin{itemize}
    \item \textbf{Direct Clustering}: Video data with shape $(B, F, Z)$ is directly reshaped into 2D with shape $(B, F \times Z)$, where $B$ is the batch size, $F$ is the number of frames, and $Z$ represents the encoded feature of each frame. Clustering is then performed on this reshaped representation, and the resulting cluster centers are reshaped back to their original 3D form.
    
    \item \textbf{Real-Video Clustering}: Based on the hypothesis that motion patterns within the same class share similar characteristics, we cluster videos using a single representative frame (spatial feature). From the resulting cluster centers, we identify the nearest real videos to retain their full temporal features. In our experiments, the first frame was selected as the representative frame for clustering.
    
    \item \textbf{Dummy-Video Clustering}: Building upon the second method, we copy the cluster center of a single frame the same number of copies as the number of video frames to create a "dummy video" without temporal variation, which serves as $m_k$.
\end{itemize}

Additionally, we experimented with other clustering methods, such as herding \cite{welling2009herding}, Gaussian Mixture Models Clustering, using their clustering centers as $m_k$. To better accommodate the 3D spatial characteristics of video data, we also investigated alternative distance metrics for clustering, such as cosine similarity, and the Frobenius norm which measures the distance between matrices, compared to the default Euclidean distance \cite{dokmanic2015euclidean}. 

\begin{figure}[ht]
  \centering
  \includegraphics[width=0.85\linewidth]{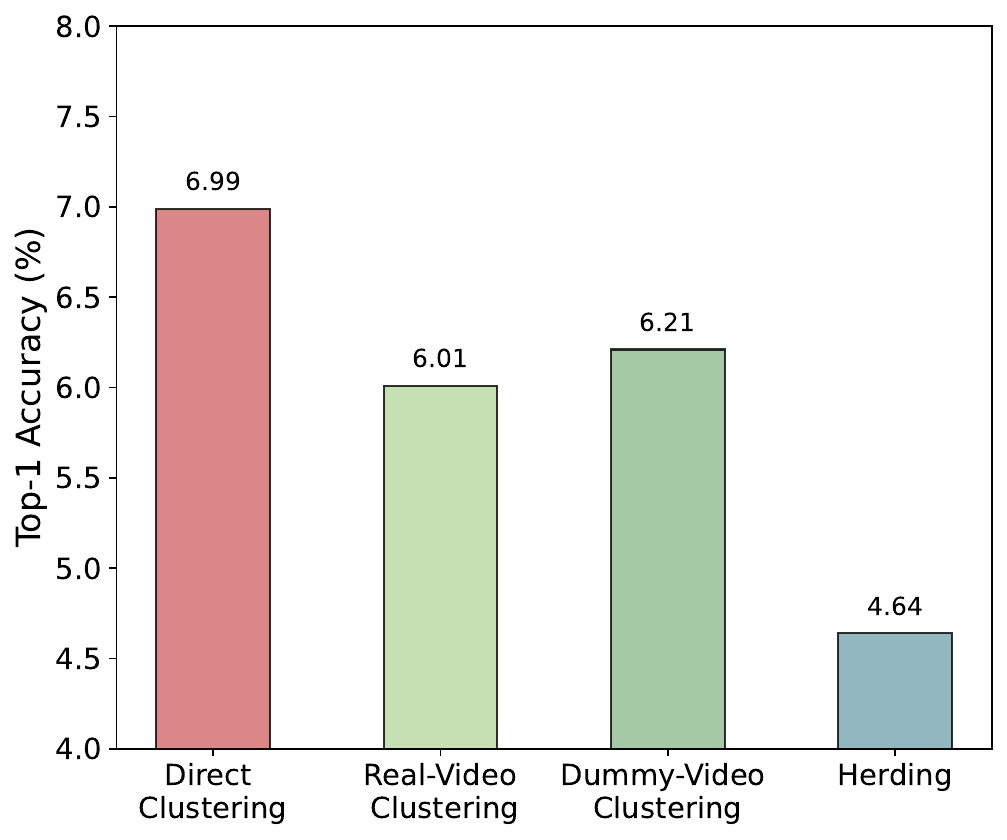} %
  \caption{Performance comparison across different clustering methods. We use the HuggingFace checkpoint for ModelScope(T2V) 1.7B for these experiments. Direct Clustering notably outperforms the other methods by a minimum of .78\%.}
  \label{fig:cluster}
\end{figure}

As shown in Figure \ref{fig:cluster}, we show the performance of the four main methods on the HMDB51 dataset under the IPC=5 setting. According to our experimental results, we found that none of these methods are as effective as direct clustering.

\section{Hyper-parameter Analysis}
\label{sec:A-6-hyperparam}

We use a MiniC3D model pre-trained on the MiniUCF full dataset to directly evaluate the videos generated under different settings, obtaining the classification accuracy.

\subsection{Guidance Strength}
\label{sec:A-7-guidance}

The results in Figure \ref{fig:para_choice} (a) shows the impact of varying the guidance strength \(\lambda\) on classification accuracy. The best performance is observed when \(\lambda = 0.1\), achieving an accuracy of 88\%. For smaller values of \(\lambda\) (e.g., 0.01), the accuracy drops significantly to 80\%, likely due to insufficient guidance. On the other hand, excessive guidance (\(\lambda = 0.5\)) results in a sharp decline to 64\%, which could be attributed to over-constrained sampling, leading to reduced diversity and poor generalization. These results highlight the importance of balancing guidance intensity for optimal performance.

\subsection{Timestep Threshold}
\label{sec:A-8-timestop}

\begin{figure}[ht]
  \centering
  \begin{subfigure}{0.48\linewidth}
    \includegraphics[width=\linewidth]{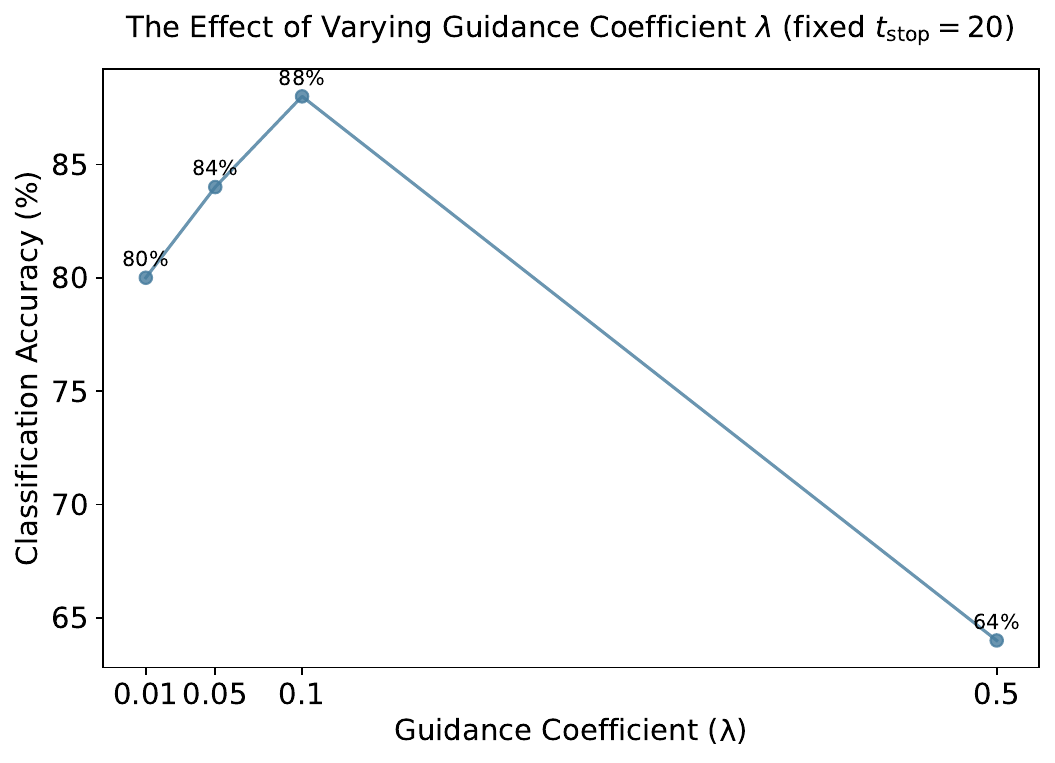} 
    \caption{}
    \label{fig:short-a}
  \end{subfigure}
  \hfill
  \begin{subfigure}{0.48\linewidth}
    \includegraphics[width=\linewidth]{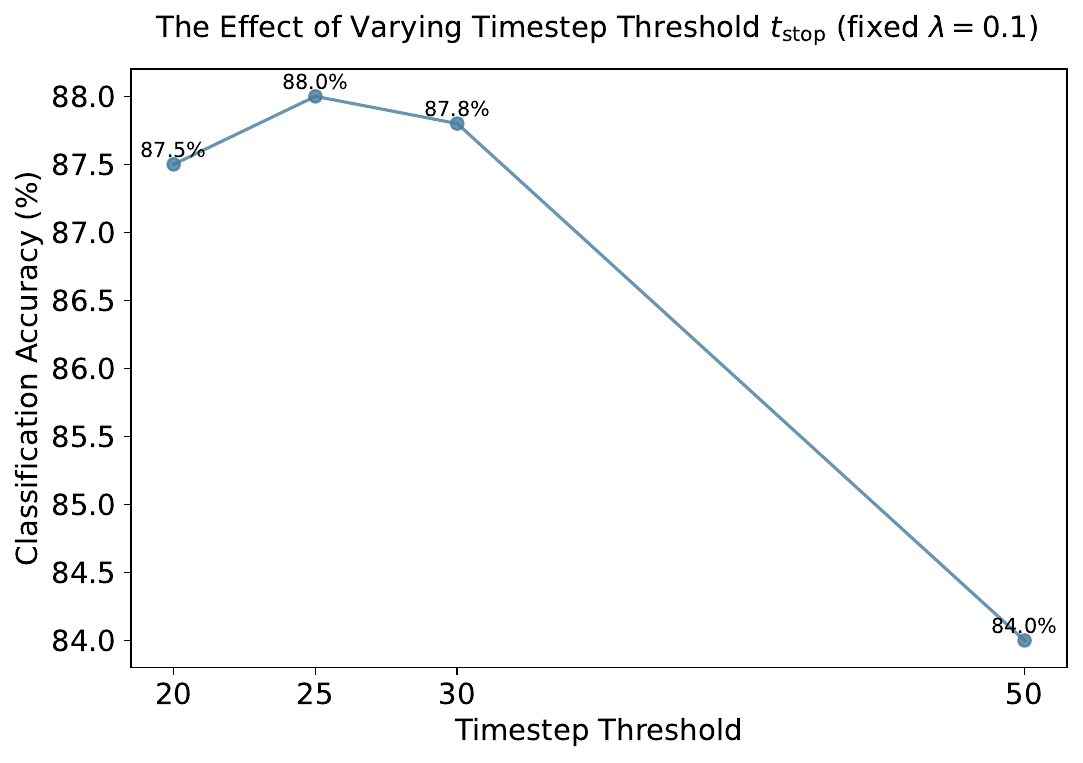} 
    \caption{}
    \label{fig:short-b}
  \end{subfigure}
  \caption{(a) Classifiation Accuracy versus Guidance Coefficient. The best is achieved with the Guidance Coefficient set to 0.1. (b) Classification Accuracy of the generated dataset versus timestep threshold $t_{stop}$. The best is achieved with a timestep threshold of 25. }
  \label{fig:para_choice}
\end{figure}

Figure \ref{fig:para_choice} (b) illustrates the relationship between the timestep threshold \(t_{\text{stop}}\) and classification accuracy. The optimal accuracy of 88\% is achieved at \(t_{\text{stop}} = 25\), indicating this threshold provides the best balance between maintaining sufficient information during sampling. A lower threshold (e.g., \(t_{\text{stop}} = 20\)) slightly reduces accuracy to 87.5\%, while a higher threshold (e.g., \(t_{\text{stop}} = 50\)) significantly degrades accuracy to 84\%. The results emphasize the need to identify an optimal number of guiding steps. 

\subsection{Multi-Video Instance Composition}
\label{sec:A-9-composition}
We investigate different Multi-Video Instance Composition strategies by varying the number of videos per group ($U$) and the frames selected per video. For instance, (2,2,2,2,2,2) indicates that each group consists of $U=6$ videos, each contributing 2 frames, forming a final 16-frame distilled video. Similarly, (4,4,4,4) means $U=4$ with 4 frames per video. Continuous selection extracts frames sequentially from each video and combines them in order. Random selection randomly samples frames from the videos, but frames from the same video remain sequentially ordered when combined. For each action class, we first generate $\text{IPC} \times U$ instances and then combine them based on the defined strategies, yielding $\text{IPC}$ final distilled videos per class.

The distilled datasets are used to train an action recognition model (MiniC3D), and the classification accuracy on the test set is reported \ref{tab:temporal_compose}. Using a grid search, we identify (4,4,4,4) as the optimal strategy, achieving the best performance.

\begin{table}[ht]
    \centering
    \caption{Analysis of group size and frame selection in Multi-Video Instance Composition on MiniUCF. Comparison of different frame compositions using continuous and random selection strategies. Experiments show that random frame selection with four videos per group, each contributing four frames, yields the best results.}

    \renewcommand{\arraystretch}{1.2}
    \resizebox{\linewidth}{!}{
    \begin{tabular}{c c c c c c} 
        \toprule
        \multirow{2}{*}{\bf Method} & \multicolumn{5}{c}{\bf MiniUCF} \\ 
        \cmidrule(lr){2-6}
         & (2,2,2,2,2,2) & (3,3,3,3,2,2) & (4,4,4,4) & (8,8) & 16 \\ 
        \midrule
        
        \textbf{Continuous} & 32.13\% & 31.70\% & 31.49\% & 31.43\% & 29.00\% \\ 
        
        \textbf{Random} & 33.10\% & 32.72\% & \textbf{33.37}\% & 29.82\% &  \\ 
        
        \bottomrule
    \end{tabular}}
    \label{tab:temporal_compose}
\end{table}

\section{Analysis of Video Diffusion Models}
\label{sec:A3}

\begin{figure}[ht]
  \centering
  \includegraphics[width=1.0\linewidth]{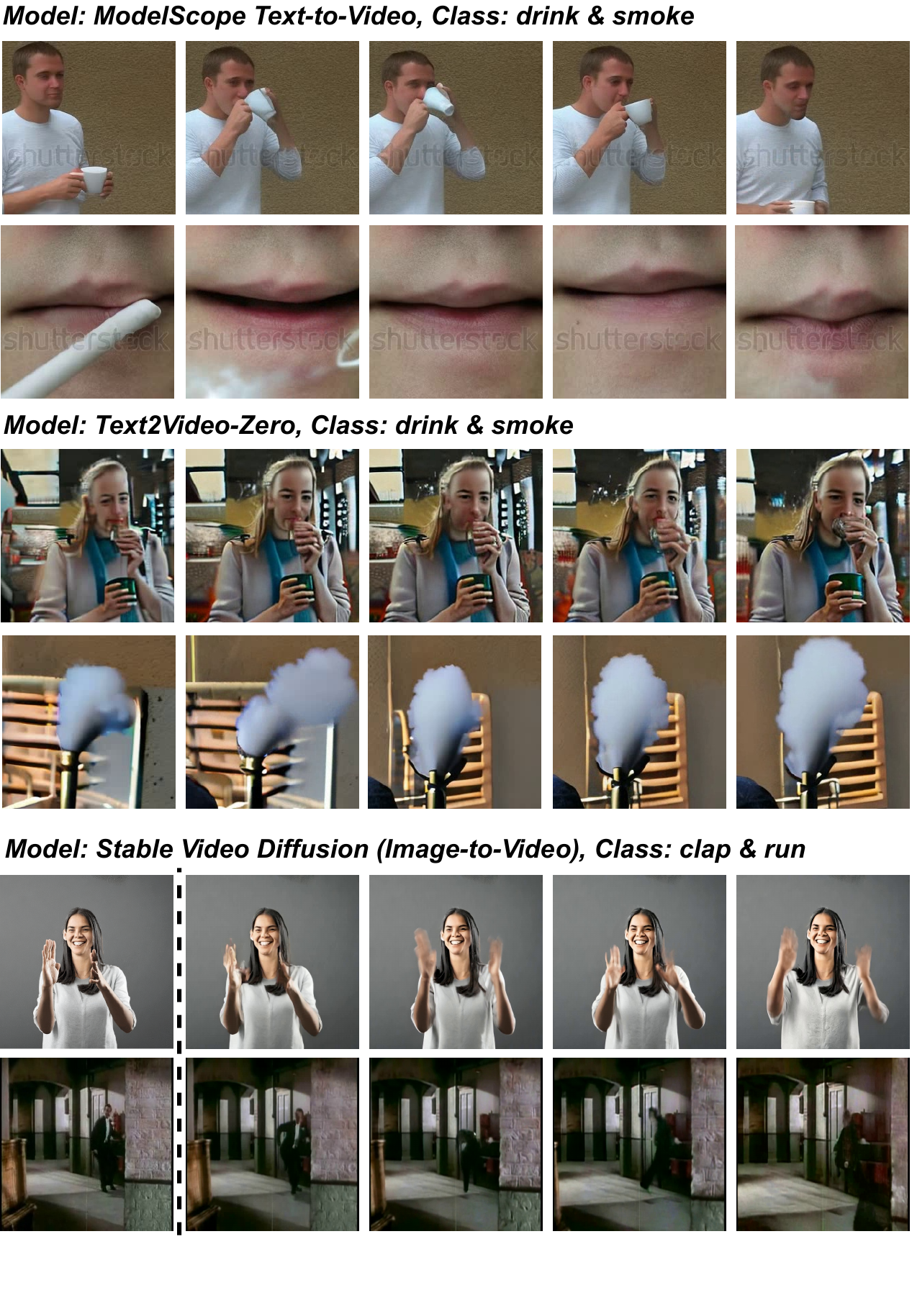} %
  \caption{Sample generated videos from ModelScope (T2V) \cite{wang2023modelscope}, Text2Video-Zero \cite{khachatryan2023text2video}, and the image-to-video model SVD \cite{blattmann2023stable}. The last two rows correspond to the SVD. Each row represents a different video, where the first image in each row serves as the image condition, and the subsequent four frames are video outputs. A non-finetuned ModelScope accurately captures the relevant spatio-temporal features of the dataset. While other models like Text2Video generate videos that are out of the dataset's distribution (in the example of smoke).}
  \label{fig:dm}
\end{figure}

In our study of video diffusion models, we evaluated two leading text-to-video generators: ModelScope (T2V) \cite{wang2023modelscope} and Text2Video-Zero \cite{khachatryan2023text2video}, both used without fine-tuning to assess baseline performance. We found ModelScope outperforms Text2Video-Zero.
Further, as shown in Figure~\ref{fig:dm}, using only image conditions without text prompts \cite{blattmann2023stable} leads to mismatched or incomplete motion generation, especially in human-centric tasks \cite{shao2025tr}. 




\section{GVD Algorithm}
\label{sec:A1}

\textit{Algorithm~\ref{alg:guided_video_diffusion}} outlines the evaluation procedure for assessing the quality of the distilled dataset. Notably, after generating the distilled dataset, we further construct the final distilled videos using the Multi-Video Instance Composition (Section~3.4). If label softening is applied during training, the loss function differs, as detailed in Section~3.5 of the paper.
\FloatBarrier

\begin{algorithm}[ht]
\scriptsize

\caption{Video Distillation via GVD}
\label{alg:guided_video_diffusion}
\textbf{Input:} \\
\hspace*{\algorithmicindent} $(\mathcal{V}, \mathcal{C})$: Training videos and their labels. \\
\hspace*{\algorithmicindent} $\mathcal{E}$: Pre-trained video encoder. \\
\hspace*{\algorithmicindent} $\mathcal{D}$: Pre-trained video decoder. \\
\hspace*{\algorithmicindent} $\epsilon_\theta$: Pre-trained time-conditional UNet. \\
\hspace*{\algorithmicindent} $K$: Number of clustering centers per class (IPC). \\
\hspace*{\algorithmicindent} $\lambda$: Guidance coefficient. \\
\hspace*{\algorithmicindent} $t_{\text{stop}}$: Timestep threshold. \\
\hspace*{\algorithmicindent} $F$: Number of video frames. \\
\textbf{Output:} $\mathcal{S}$: Distilled video dataset. \\

\begin{algorithmic}[1]
\State \textbf{Step 1: Encode Videos into Latent Space.}
\State $Z = \mathcal{E}(\mathcal{V})$ \Comment{Encode training videos into latent space.}

\State \textbf{Step 2: Generate Diversity-Feature Guiding.}
\For{each class $c \in \mathcal{C}$}
    \State Apply KMeans clustering on $Z_c$ to obtain cluster centers $\{m_{k, c}\}_{k=1}^K$
\EndFor

\State \textbf{Step 3: Guide Denoising Process.}
\For{each class $c \in \mathcal{C}$ and each cluster center $m_{k, c}$}
    \State Initialize $Z_T \sim \mathcal{N}(0, I)$ \Comment{Start diffusion process with random noise.}
    \For{$\hat{t} = T$ to $1$}
        \For{$f = 1$ to $F$}
            \State Compute frame-wise guidance coefficient:
            \[
            \lambda_f = \lambda \cdot \left( 1 - \frac{f}{F} \right)
            \]
            \State Compute guidance term:
            \[
            g_{\hat{t}, f} = m_{k, c} - \hat{x}_{0, \hat{t}, f}
            \]
            \State Update noise prediction:
            \[
            \epsilon'_{\hat{t}, f} = 
            \begin{cases} 
            \epsilon_\theta(Z_{\hat{t}, f}, c, \hat{t}) - \lambda_f \cdot \sqrt{1 - \bar{\alpha}_{\hat{t}}} \cdot g_{\hat{t}, f}, & \hat{t} < t_{\text{stop}} \\
            \epsilon_\theta(Z_{\hat{t}, f}, c, \hat{t}), & \hat{t} \geq t_{\text{stop}}
            \end{cases}
            \]
            \State Update latent representation:
            \[
            Z_{\hat{t}-1, f} = \sqrt{\bar{\alpha}_{\hat{t}-1}} \cdot \hat{x}_{0, \hat{t}} + \sqrt{1 - \bar{\alpha}_{\hat{t}-1}} \cdot \epsilon'_{\hat{t}, f}
            \]
        \EndFor
    \EndFor
    \State Decode latent representation:
    \[
    v^{pr} = \mathcal{D}(\{Z_{0, f}\}_{f=1}^F)
    \]
    \State Add $v^{pr}$ to distilled dataset $\mathcal{S}$.
\EndFor

\State \textbf{Output:} $\mathcal{S}$ \Comment{Distilled video dataset.}
\end{algorithmic}
\end{algorithm}


\FloatBarrier

\section{Qualitative Results}
\label{sec:A-9-qualitative}
\begin{figure*}[t]
  \centering
  \includegraphics[width=0.9\linewidth]{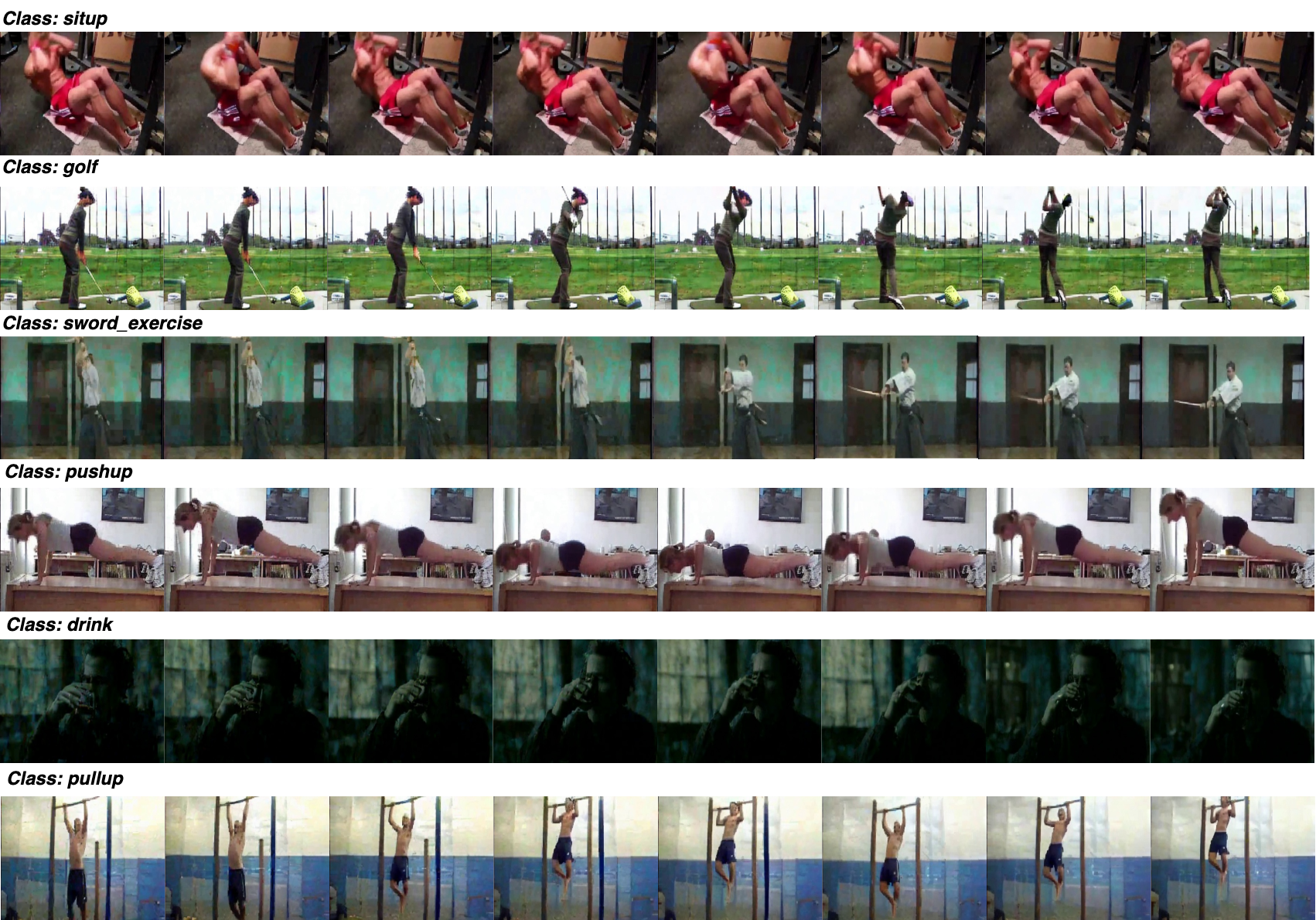}
  \caption{Qualitative results from HMDB51 \cite{kuehne2011hmdb} with IPC=10.}
  \label{fig:HMDB51 DEMO}

  \vspace{0.5cm}

  \includegraphics[width=0.9\linewidth]{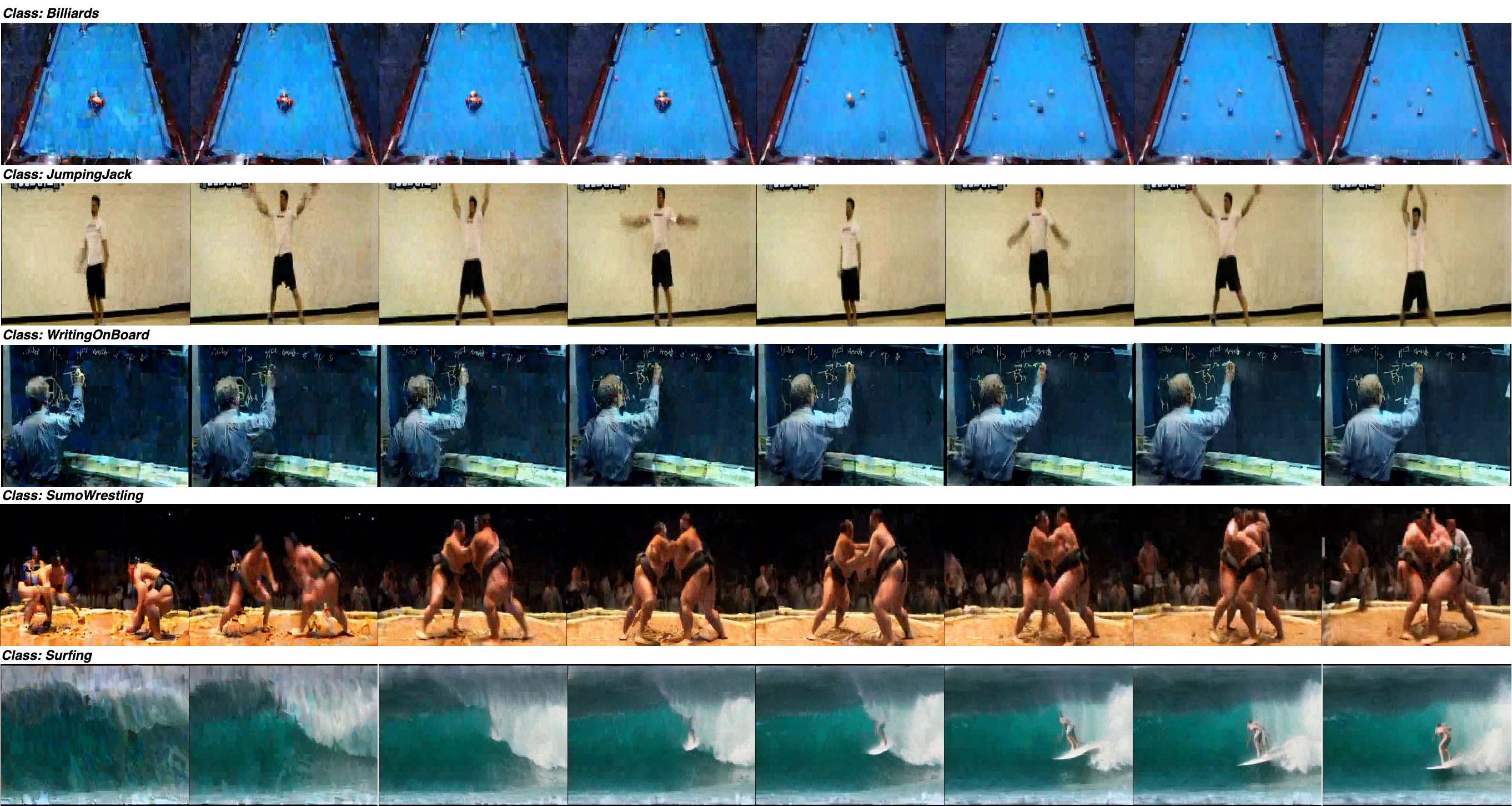}
  \caption{Qualitative results from MiniUCF \cite{soomro2012ucf101} with IPC=10.}
  \label{fig:UCF101 DEMO}
\end{figure*}

In this section, we present some generated videos from our method. We show results for datasets HMDB51\cite{kuehne2011hmdb}, MiniUCF\cite{soomro2012ucf101} in Figures \ref{fig:HMDB51 DEMO} and \ref{fig:UCF101 DEMO}, respectively. These examples demonstrate that our method can effectively generate videos across many classes and datasets. These results are obtained without the Multi-Video Instance Composition, providing a clearer view of the direct impact of our guiding mechanism for dataset distillation.

\end{document}